\documentclass[journal]{IEEEtran}

\usepackage{times}
\usepackage{epsfig}
\usepackage{graphicx}
\usepackage{amsmath}
\usepackage{amssymb}
\usepackage{algorithmic}
\usepackage{multirow}
\usepackage[table,xcdraw]{xcolor}

\usepackage[breaklinks=true,bookmarks=false]{hyperref}

\usepackage{caption}

\graphicspath{{Figs/}}
\DeclareMathOperator*{\argmin}{argmin}

\newcommand{\etal}{\textit{et al.}}
\begin{document}

\title{Deblurring Face Images using Uncertainty Guided\\ Multi-Stream Semantic Networks}

\author{Rajeev Yasarla,~\IEEEmembership{Student Member,~IEEE}, Federico Perazzi,~\IEEEmembership{Member,~IEEE} and 
	Vishal~M.~Patel,~\IEEEmembership{Senior Member,~IEEE}
	\thanks{Rajeev Yasarla is with the Whiting School of Engineering, Johns Hopkins University, 3400 North Charles Street, Baltimore, MD 21218-2608, e-mail: ryasarl1@jhu.edu}
		\thanks{Federico Perazzi is with Adobe Inc., e-mail: perazzi@adobe.com}
	\thanks{Vishal M. Patel is with the Whiting School of Engineering, Johns Hopkins University, e-mail: vpatel36@jhu.edu}
	\thanks{Manuscript received...}}

\maketitle

\begin{abstract}
	We propose a novel multi-stream architecture and training methodology that exploits semantic labels for facial image deblurring. The proposed Uncertainty Guided Multi-Stream Semantic Network (UMSN) processes regions belonging to each semantic class independently and learns to combine their outputs into the final deblurred result. Pixel-wise semantic labels are obtained using a segmentation network. A predicted confidence measure is used during training to guide the network towards the challenging regions of the human face such as the eyes and nose. The entire network is trained in an end-to-end fashion.  Comprehensive experiments on three different face datasets demonstrate that the proposed method achieves significant improvements over the recent state-of-the-art face deblurring methods. Code is available at: \url{https://github.com/rajeevyasarla/UMSN-Face-Deblurring}
\end{abstract}


\IEEEpeerreviewmaketitle

\section{Introduction}

Image deblurring entails the recovery of an unknown true image from a blurry image.  Similar to the other image enhancement tasks, image deblurring is experiencing a renaissance as a result of convolutional networks (CNNs) establishing themselves as powerful generative models.  Image deblurring is an ill-posed problem and therefore it is crucial to leverage additional properties of the data to successfully recover the lost facial details in the deblurred image. Priors such as sparsity \cite{Fergus06removingcamera,Schuler2013, ShearDeconv}, manifold \cite{ManifoldDeblur}, low-rank \cite{Ren2016} and patch similarity \cite{patchdeblur_iccp2013} have been used in the literature to obtain a regularized solution.  In recent years, deep learning-based methods have also gained some traction   \cite{Nimisha17,ayan2016,Nah_2017_CVPR,Zhang_2018_CVPR}.

\begin{figure}[htp!]
	\centering
	\includegraphics[width=0.115\textwidth]{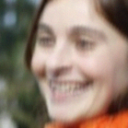}
	\includegraphics[width=0.115\textwidth]{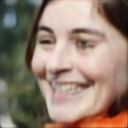} 
	\includegraphics[width=0.115\textwidth]{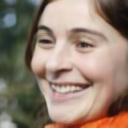}
	\includegraphics[width=0.115\textwidth]{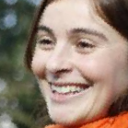}\\
	\vskip2pt
	\includegraphics[width=0.115\textwidth]{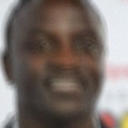}
	\includegraphics[width=0.115\textwidth]{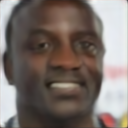} 
	\includegraphics[width=0.115\textwidth]{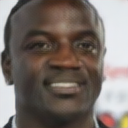}
	\includegraphics[width=0.115\textwidth]{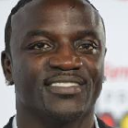}\\
	\vskip2pt
	\includegraphics[width=0.115\textwidth,height= 0.115\textwidth]{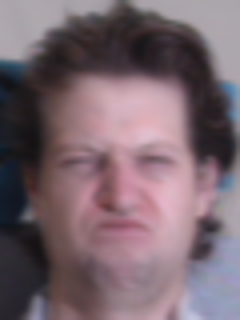}
	\includegraphics[width=0.115\textwidth,height= 0.115\textwidth]{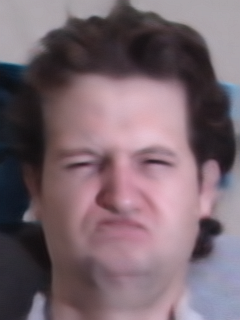} 
	\includegraphics[width=0.115\textwidth,height= 0.115\textwidth]{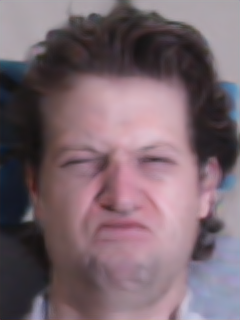}
	\includegraphics[width=0.115\textwidth,height= 0.115\textwidth]{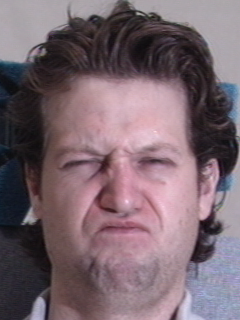}\\
	(a)\hskip50pt(b)\hskip50pt(c)\hskip50pt(d)\\
	\caption{Sample deblurring results: (a) Blurry image; (b) results corresponding to Shen \etal~\cite{ZiyiDeep} and to Song \etal\cite{song-ijcv19-FHD} (last row); (c) Results corresponding to our proposed Uncertainty Guided Multi-Stream Semantic Network (UMSN); (d) ground-truth. Our approach recovers more details and better preserves fine structures like eyes and hair.
	}
	\label{Fig:exp1}
\end{figure}

The inherent semantic structure of natural images such as faces is an important information that can be exploited to improve the deblurring results. Few techniques \cite{song-ijcv19-FHD,ZiyiDeep} make use of such prior information in the form of semantic labels. These methods do not account for the class imbalance of semantic maps corresponding to faces. Interior parts of a face like eyes, nose, and mouth are less represented as compared to face skin, hair and background labels. Depending on the pose of the face, some of the interior parts may even disappear. Without re-weighting the importance of less represented semantic regions, the method proposed by Shen \etal\cite{ZiyiDeep} fails to reconstruct the eyes and the mouth regions as shown in Fig. \ref{Fig:exp1}. Similar observations can also be made regarding the method proposed in \cite{song-ijcv19-FHD} (Fig. ~\ref{Fig:exp1}) which uses semantic priors to obtain the intermediate outputs and then performs post-processing using k-nearest neighbor algorithm.

\begin{figure*}[htp!]
	\begin{center}
		\centering
		\includegraphics[width=\textwidth]{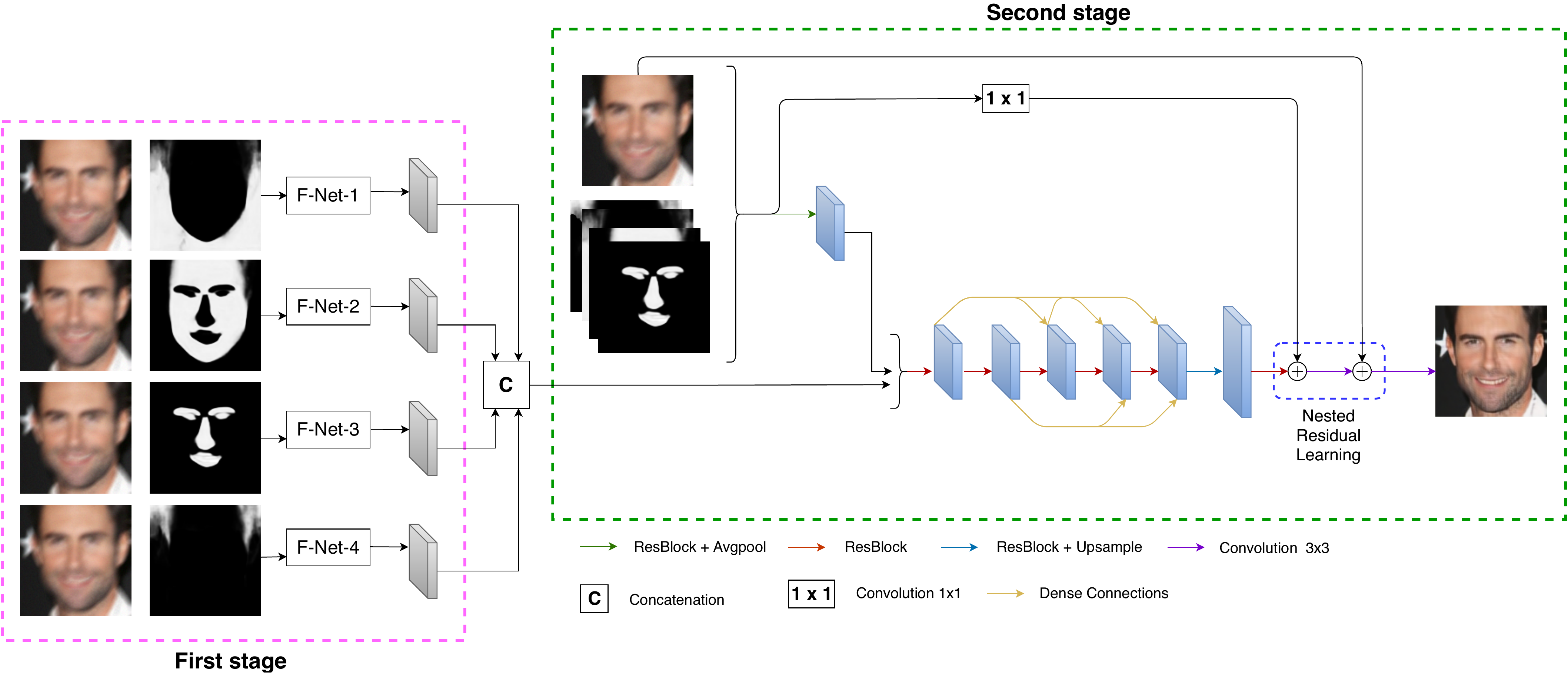}\\
		\vskip 5pt
		\caption{An overview of the proposed UMSN network. First stage semantic networks consist of F-Net-$i$. Second stage is constructed using the base network (B-Net), where outputs of the F-Net-$i$'s are concatenated with the output of the first ResBlock layer in B-Net.}
		\label{Fig:UMSN}
	\end{center}
\end{figure*}

To address the imbalance of different semantic classes, we propose a novel CNN architecture, called Uncertainty guided Multi-stream Semantic Networks (UMSN), which learns class-specific features independently and combine them to deblur the whole face image. Class-specific features are learned by sub-networks trained to reconstruct a single semantic class. We use nested residual learning paths to improve the propagation of semantic features. Additionally, we propose a class-based confidence measure to train the network. The confidence measure describes how well the network is likely to deblur each semantic class. This measure is incorporated in the loss to train the network. We evaluate the proposed network by conducting experiments on three face datasets - Helen  \cite{helen2012}, CelebA  \cite{liu2015faceattributes} and PuBFig \cite{song-ijcv19-FHD}. Extensive experiments demonstrate the effectiveness of our approach compared to a number of well established face deblurring methods as well as other competitive approaches. Fig. \ref{Fig:exp1} shows sample results from our UMSN network, where one can clearly see that UMSN is able to provide  better results as compared to the state-of-the-art techniques \cite{ZiyiDeep,song-ijcv19-FHD}.   An ablation study is also conducted to demonstrate the effectiveness of different parts of the proposed network. To summarize, this paper makes the following contributions:
\begin{itemize}
	\item A novel multi-stream architecture called, Uncertainty Guided Multi-Stream Semantic (UMSN), is proposed which learns class-specific features and uses them to reconstruct the final deblurred image.
	\item We propose a novel method of computing confidence measure for the reconstruction of every class in the deblurred image, which is to rebalance the importance of semantic classes during training.
\end{itemize}
Rest of the paper is organized as follows. In Section~\ref{sec:bacground}, we review a few related works. Details of the proposed  method are given in Section~\ref{sec:method}. Experimental results are presented in Section~\ref{sec:results}, and finally, Section~\ref{sec:con} concludes the paper with a brief summary.

\section{Background and Related Work}\label{sec:bacground} 
Single image deblurring techniques can be divided into two main categories - blind and non-blind.  In the non-blind deblurring problem, the blur kernel is assumed to be known while in the blind deblurring problem, the blur kernel needs to be estimated.  Since the proposed approach does not assume any knowledge of the blur kernel, it is a blind image deblurring approach.  Hence, in this section, we review some recent blind image deblurring techniques proposed in the literature.

Classical image deblurring methods estimate the blur kernel given a blurry image and then apply deconvolution to get the deblurred image. To calculate the blur kernel some techniques assume prior information about the image, and formulate maximum a-posertior (MAP) to obtain the deblurred image.  Different priors such sparsity, $L_0$ gradient prior, patch prior, manifold Prior, and low-rank have been proposed to obtain regularized reconstructions  \cite{Xu2013,Fergus06removingcamera,Krishnan2011,Ren2016,Schuler2013,Pan_2016_CVPR,patchdeblur_iccp2013,ManifoldDeblur,ShearDeconv}.  Chakrabarti \etal\cite{ayan2016} estimate the Fourier coefficients of the blur kernel to deblur the image. Nimisha \etal\cite{Nimisha17} learn the latent features of the blurry images using the latent features of the clean images to estimate the deblurred image. These CNN-based methods do not perform well compared to the state-of-the-art MAP-based methods for large motion kernels.  Recent non-blind image deblurring methods like \cite{Kruse2017,Schmidt2013,Schuler2013,Schmidt2014,vasu2018non} assume and use some knowledge about the blur kernel. Given the latent blurry input, Subeesh \etal\cite{vasu2018non} estimate multiple latent images  corresponding to different prior strengths to estimate the final deblurred image.  CNN-based techniques like \cite{Schuler_NIPS14} and \cite{Xu2014} estimate the blur kernel and address dynamic deblurring. Another recent image deblurring method based on Feature Pyramid Network and conditional  generative adversarial network (GAN) was recently proposed by Kupyn \etal in \cite{kupyn2019deblurgan}.

The usage of semantic information for image restoration is relatively unexplored.  While semantic information has been used for different object classes \cite{TsaiSLSY16,EigenKF13,SvobodaHMZ16,PanHSY14}, a substantial body of literature has focused their attention to human faces \cite{BakerK02,ZiyiDeep,PanECCV14,JoshiMAK10}. Pan \etal \cite{PanECCV14} extract the edges of face parts and estimate exemplar face images, which are further used as the global prior to estimate the blur kernel.  This approach is complex and computationally expensive in estimating the blur kernel.  Recently, Shen \etal \cite{ZiyiDeep} proposed to use the semantic maps of a face to deblur the image. Furthermore, they introduced the content loss to improve the quality of eyes, nose and mouth regions of the face. Bulat \textit{et al.}~\cite{bulat2018super} proposed Super-FAN architecture which is an end-to-end network that outputs high-resolution of face, along with heatmap for face alignment. Chen \textit{et al.}~\cite{chen2018gated} proposed FSRGAN architecture which consists of coarse and fine Super-Resolution (SR) networks where they estimate coarse estimate of high-resolution output image, follwed by fine estimate using Fine SR Network. Yu \textit{et al.}~\cite{yu2018face} propose a multi-task network to upsample LR image where they predict face structure along with super-resolving LR face image. Ren \etal~\cite{ren2019face} proposed a face video deblurring method by predicting facial structure and identity from the blurry face using a deep network that generates a textured 3D face from the video. Textured 3D face mask can be generated accurately only from videos.  This method is not beneficial if we have a single face image to deblur.  Lu \etal~\cite{lu2019unsupervised} proposed a domain-specific single face image deblurring method by disentangling the content information in an unsupervised fashion using the KL-divergence. In contrast to these methods, we learn a multi-stream network which reconstructs the deblurred images corresponding to different classes in a facial semantic map. Furthermore, we propose a new loss to train the network.

\section{Proposed Method}\label{sec:method}
A blurry face image $y$ can be modeled as the convolution of a clean image $x$ with a blur kernel $k$, as $$y = k * x + \eta,$$ where $*$ denotes the convolution operation and $\eta$ is noise. Given $y$, in blind deblurring, our objective is to estimate the underlying clean face image $x$. We group 11 semantic face labels into 4 classes as follows: $m_1 = \{background\}$, $m_2= \{face\;skin\}$, $m_3=\{left\;eyebrow,\;right\,\:eyebrow,\;left\;eye,\;right\;eye,\newline \;nose,\;upper\;lip,\;lower\;lip,\;teeth\}$ and $m_4 = \{hair\}$. Thus, the semantic class mask of a clean image $x$ is the union $m=m_1 \cup m_2 \cup m_3 \cup m_4$. Similarly we define the semantic class masks of a blurry image $\hat{m}$.

\begin{figure}[htp!]
	\centering
	\includegraphics[width=0.48\textwidth]{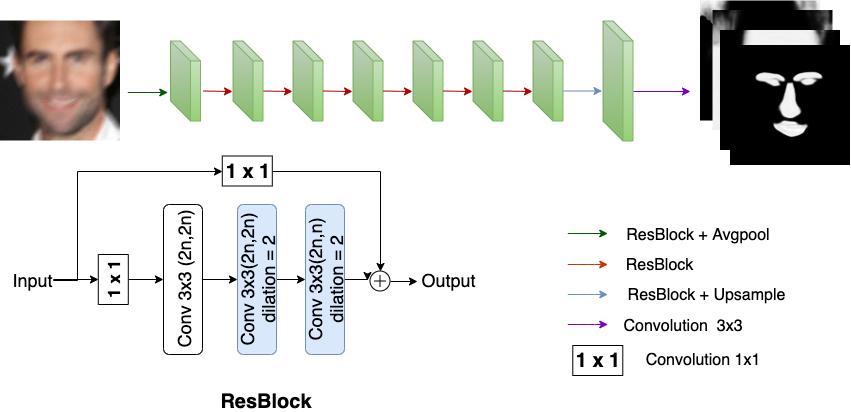}\\
	\caption{An overview of the segmentation network. Conv~$l\times~l$~($p,q$) contains instance normalization \cite{Instance2016}, Rectified Linear Unit (ReLU), Conv ($l \times l$) - convolutional layer with kernel of size $l \times l$, where $p$ and $q$ are the number of input and output channels respectively}
	\label{Fig:segmentation}
\end{figure}

Semantic class masks for blurry image, $\hat{m}$ are generated using the semantic segmentation network (S-Net) (Fig.~\ref{Fig:segmentation}), and given together with the blurry image as input to the deblurring network, UMSN. This is important in face deblurring as some parts like face skin and hair are easy to reconstruct, while face parts like eyes, nose, and mouth are difficult to reconstruct and require special attention while deblurring a face image. This is mainly due to the fact that parts like eyes, nose and mouth are small in size and contain high frequency elements compared to the other components.  Different from \cite{PanECCV14} that uses edge information and \cite{ZiyiDeep} that feed the semantic map to a single-stream deblurring network, we address this problem by proposing a multi-stream semantic network, in which individual branches F-net-$i$ learn to reconstruct different parts of the face image separately. Fig.~\ref{Fig:UMSN} gives an overview of the proposed UMSN method.

As can be seen from Fig.~\ref{Fig:UMSN}, the proposed UMSN network consists of two stages.  We generate the semantic class maps, $\hat{m}$, of a blurry face image using the S-Net network.  The semantic maps are used as the global masks to guide each stream of the first stage network.  These semantic class maps $\hat{m}$ are further used to learn class specific residual feature maps with nested residual learning paths (NRL). In the first stage of our network, the weights are learned to deblur the corresponding class of the face image.  In the second stage of the network, the outputs from the first stage are fused  to learn the residual maps that are added to the blurry image to obtain the final deblurred image.  We train the proposed network with a confidence guided class-based loss.

\subsection{Semantic Segmentation Network (S-Net)}
The semantic class maps $\hat{m}_{i}$ of a face, are extracted using the S-Net network as shown in Fig. \ref{Fig:segmentation}. We use residual blocks (ResBlock) as our building module for the segmentation network. A ResBlock consist of a $1 \times 1$ convolution layer, a $3 \times 3$ convolution layer and two $3 \times 3$ convolution layers with dilation factor of 2 as shown in Fig. \ref{Fig:building_blocks}.

\subsection{Base Network (B-Net)}
\begin{figure}[htp!]
	\centering
	\includegraphics[width=0.48\textwidth]{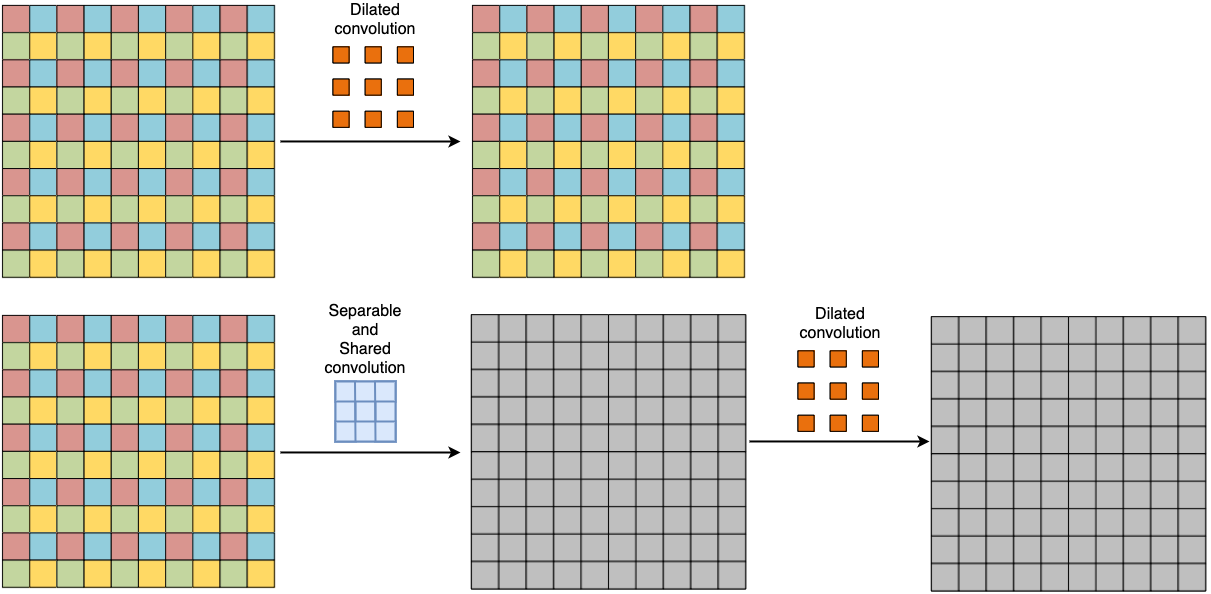}\\ \vskip5pt
	\includegraphics[width=0.48\textwidth]{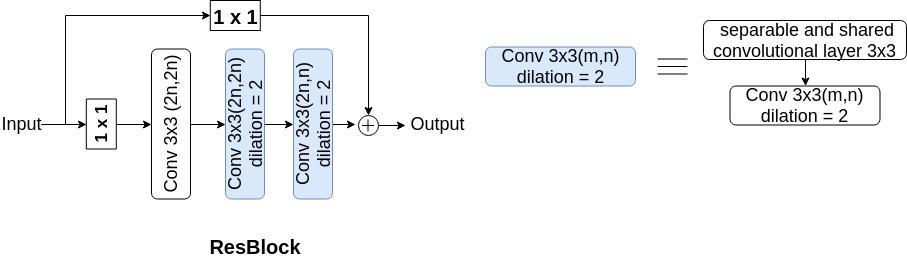}\\
	\caption{First row shows an example of gridding effects caused by the dilated convolutions.  For instance, a red pixel in the output feature map will be a function of the corresponding red pixels in the input feature map. This causes gridding artifacts. Second row shows the benefit of using separable and shared convolution layer, where every pixel in the output feature map is a function of every pixel in the corresponding neighborhood of the input feature map. Third row is an overview of the ResBlock. Conv $l\times l$($p,q$) contains Instance Normalization \cite{Instance2016}, ReLU - Rectified Linear Units, convolutional layer with kernel of size $l \times l$, where $p$ and $q$ are number of input and output channels respectively. In the right side of the figure, we show smoothed dilation convolutions introduced in ResBlock which is similar to \cite{wang2018smoothed}}
	\label{Fig:building_blocks}
\end{figure}

We construct our base network using a combination of UNet~\cite{Authors15b} and DenseNet~\cite{huang2017densely} architectures with the ResBlock as our basic building block. To increase the receptive field size, we introduce smoothed dilation convolutions in the ResBlock as shown in Fig.~\ref{Fig:building_blocks}. B-Net is a sequence of eight ResBlocks similar to the first stage semantic network as shown in Fig.~\ref{Fig:F-Net}. Note that all convolutional layers are densely connected \cite{huang2017densely}. We follow residual-based learning in estimating the deblurred image for our base network as shown in Fig. \ref{Fig:F-Net}.

\subsection{UMSN Network}
\begin{figure}[h!]
	\centering
	\includegraphics[width=0.48\textwidth]{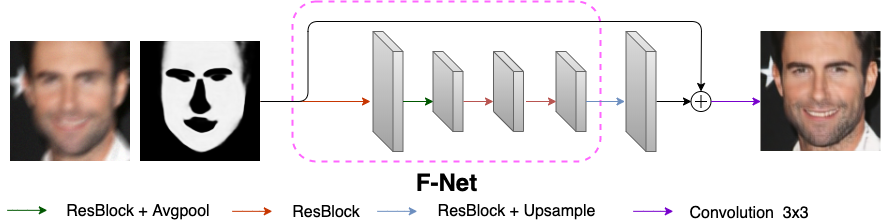}\\
	\caption{An overview of the first stage semantic network. We define the set of all convolution layers of the first stage network excluding the last ResBlock and Conv$3\times 3$ as F-Net.}
	\label{Fig:F-Net}
\end{figure}

The UMSN network is a two-stage network.  The first stage network is designed to obtain deblurred outputs from the semantic class-wise blurry inputs.  These outputs are further processed by the second stage network to obtain the final deblurred image.  The first stage semantic network contains a sequence of five ResBlocks with residual connections, as shown in Fig.~\ref{Fig:F-Net}. We call the set of all convolution layers of the first stage network excluding the last ResBlock and Conv$3\times 3$ as F-Net.

The blurry image $y$ and the semantic masks $\hat{m}_i$ are fed to F-Net-$i$ to obtain the corresponding class-specific deblurred features which are concatenated with the output of the first layer (ResBlock-Avgpool) in Base Network(B-Net) for constructing the UMSN network. We also use the Nested Residual Learning (NRL) in UMSN network where class specific residual feature maps are learned and further used in estimating the residual feature maps that are added to the blurry image for obtaining the final output. For example, we can observe a residual connection between the last layer of UMSN and class-specific feature maps obtained from Conv $1\times 1$ using $y$ and $\hat{m}$.  Output of this residual connection is further processed as the input to the residual connection with the input blurry image, $y$. In this way, we define our NRL and obtain the final deblurred image. We propose a class-based loss function to train the UMSN network.

\subsection{Loss for UMSN}
The network parameters $\Theta$ are learned by minimizing a loss $\mathcal{L}$ as follows,
\begin{equation}\label{eq:loss}
\hat{\Theta} = \argmin_{\Theta} \mathcal{L}(f_\Theta(y,\hat{m}),x) = \argmin_{\Theta} \mathcal{L}(\hat{x},x),
\end{equation}
where $f_\Theta(.)$ represents the UMSN network, $\hat{x}$ is the deblurred result, $\hat{m}$ is the semantic map obtained from S-Net. We define the reconstruction loss as $\mathcal{L}=\|x-\hat{x}\|_{1}$. A face image can be expressed as the sum of masked images using the semantic maps as $$x = \sum_{i=1}^{M} m_i \odot x,$$ where $\odot$ is the element-wise multiplication and $M$ is the total number of semantic maps. As the masks are independent of one another, Eq.~\eqref{eq:loss} can be re-written as,
\begin{equation}
\hat{\Theta} =  \argmin_{\Theta} \sum_{i=1}^{M}\mathcal{L}(m_i \odot \hat{x}, m_i \odot x).
\end{equation}
In other words, the loss is calculated for every class independently and summed up in order to obtain the overall loss as follows,

\begin{equation} \label{eq:loss_tmp}
\mathcal{L}(\hat{x},x) = \sum_{i=1}^{M}\mathcal{L}(m_i \odot \hat{x}, m_i \odot x).
\end{equation}

\subsection{Uncertainty Guidance} 
\begin{figure}[htp!]
	\centering
	\includegraphics[width=8cm]{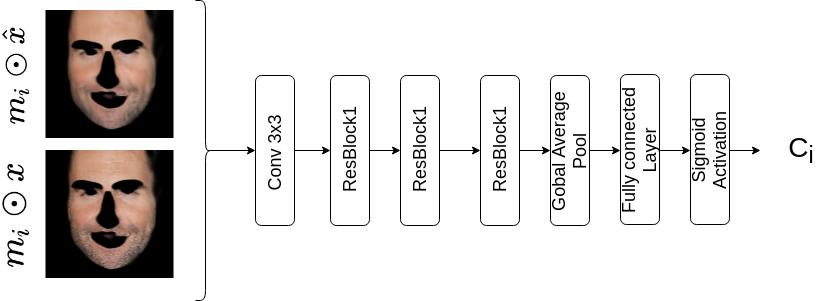}\\
	\caption{An overview of the Confidence Network (CN). $x$ is ground truth image. $\hat{x}$ is deblurred image obtained from UMSN. $m$ semantic maps of $x$}
	\label{Fig:Confidence}
	
\end{figure}

We introduce a confidence measure for every class and use it to re-weight the contribution of the loss from each class to the total loss. By introducing a confidence measure and re-weighting the loss, we benefit in two ways. If the network is giving less importance to a particular class by not learning appropriate features of it, then the Confidence Network (CN) helps UMSN to learn those class specific features by estimating low confidence values and higher gradients for those classes through the CN network. Additionally, by re-weighting the contribution of loss from each class, it counters for the imbalances in the error estimation from different classes. The loss function can be written as,
\begin{equation}
\mathcal{L}_c(\hat{x},x) = \sum_{i=1}^{M} C_i  \mathcal{L}(m_i \odot \hat{x}, m_i \odot x) - \lambda \log (C_i),
\end{equation}
where $\log (C_i)$ acts as a regularizer that prevents the value of $C_i$ going to zero and $\lambda$ is a constant. We estimate the confidence measure $C_i$ for each class by passing $m_i \odot \hat{x}, m_i \odot x$ as inputs to CN as shown in Fig.~\ref{Fig:Confidence}. $C_i$ represents how confident UMSN is in deblurring the $i^{th}$ class components of the face image. Note that, $C_i$($\in[0,1]$), confidence measure is used only in the loss function while training the weights of UMSN, and it is not used (or estimated) during inference.

Inspired by the benefits of the perceptual loss in style transfer \cite{Johnson2016Perceptual,zhang2017multistyle} and image super-resolution \cite{Ledig2017PhotoRealisticSI}, we use it to train our network.   Let $\Phi(.)$ denote the features obtained using the VGG-Face model~\cite{Parkhi15}, then the perceptual loss is defined as follows,
\begin{equation}
\mathcal{L}_p = \lVert \Phi(\hat{x})-\Phi(x)\rVert_2^{2}.
\end{equation}
The features from layer $relu1\_2$ of a pretrained VGG-Face network \cite{Parkhi15} are used to compute the perceptual loss.  The total loss used to train UMSN is as follows,
\begin{equation}
\mathcal{L}_{total} = \mathcal{L}_c + \lambda_1 \mathcal{L}_p,
\end{equation}
where $\lambda_1$ is a constant.

	\begin{figure*}[htp!]
		\centering
		\includegraphics[width=0.15\textwidth]{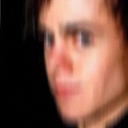}
		\includegraphics[width=0.15\textwidth]{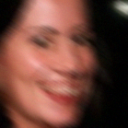}
		\includegraphics[width=0.15\textwidth]{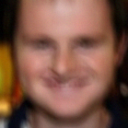}
		\includegraphics[width=0.15\textwidth]{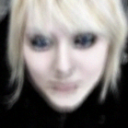}
		\includegraphics[width=0.15\textwidth]{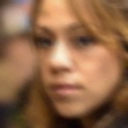}
		\includegraphics[width=0.15\textwidth]{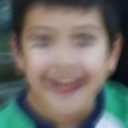}\\ \vskip2pt
		\includegraphics[width=0.15\textwidth]{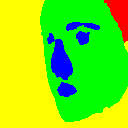}
		\includegraphics[width=0.15\textwidth]{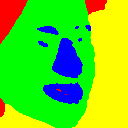}
		\includegraphics[width=0.15\textwidth]{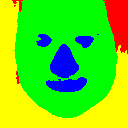}
		\includegraphics[width=0.15\textwidth]{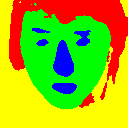}
		\includegraphics[width=0.15\textwidth]{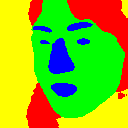}
		\includegraphics[width=0.15\textwidth]{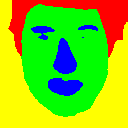}
		\\ \vskip2pt
		\includegraphics[width=0.15\textwidth]{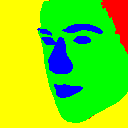}
		\includegraphics[width=0.15\textwidth]{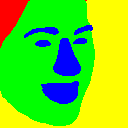}
		\includegraphics[width=0.15\textwidth]{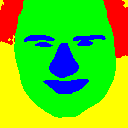}
		\includegraphics[width=0.15\textwidth]{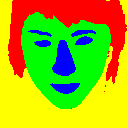}
		\includegraphics[width=0.15\textwidth]{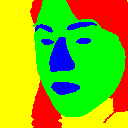}
		\includegraphics[width=0.15\textwidth]{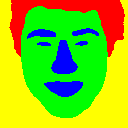}
		\caption{Semantic maps generated by S-Net on blurry images from the Helen dataset. First row contains Blurry images. Second rows consists of the corresponding Semantic masks obtained from S-Net trained on the clean images. Third row consists of the corresponding Semantic masks obtained from S-Net fine tuned with the blurry images.}
		\label{Fig:Seg}
		
	\end{figure*}

\section{Experimental Results}\label{sec:results}
We evaluate the results using the images provided by the authors of \cite{ZiyiDeep} which consists of $8000$ blurry images generated using the Helen dataset \cite{helen2012}, and $8000$ blurry images generated using the CelebA dataset \cite{liu2015faceattributes}\footnote{We follow the same training and testing protocols defined by \cite{ZiyiDeep}.}. Furthermore, we test our network on a test dataset called PubFig, provided by the authors of \cite{song-ijcv19-FHD} which contains 192 blurry images. We quantitatively evaluate the results using the Peak-Signal-to-Noise Ratio (PSNR) and the Structural Similarity index (SSIM) \cite{SSIM}.

\setlength{\tabcolsep}{5pt}
\def\arraystretch{1.3}
\begin{table}[t!]
	\resizebox{0.48\textwidth}{!}{
		\begin{tabular}{c|c|c|c|c|c|c|c|c|c}
			\hline
			\multirow{2}{*}{} & \multirow{2}{*}{Method} & \multicolumn{2}{c|}{class-1} & \multicolumn{2}{c|}{class-2} & \multicolumn{2}{c|}{class-3} & \multicolumn{2}{c}{class-4} \\ \cline{3-10}
			&  & PSNR  & SSIM & PSNR & SSIM & PSNR          & SSIM & PSNR & SSIM \\ \hline
			\multirow{2}{*}{C}& Shen \etal& 27.86 & 0.88 & 30.27 & 0.93 & 33.33         & 0.92 & 31.79  & 0.91 \\ \cline{2-10}
			& UMSN& 29.27 & 0.91 & 30.75 & 0.95 & 33.83 & 0.96 & 33.10 & 0.94    \\ \hline
			\multirow{2}{*}{H}& Shen \etal & 29.13 & 0.85 & 31.20 & 0.88 & 35.13  & 0.92 & 33.96  & 0.88 \\ \cline{2-10}
			& UMSN  & 30.83 & 0.87 & 31.76 & 0.90 & 35.67 & 0.93 & 34.81 & 0.91    \\ \hline
		\end{tabular}
	}
	\vspace{3pt}
	\caption{PSNR and SSIM values for each deblurred semantic class on the CelebA dataset (C) and Hellen dataset (H).
		\label{Table:class}}
\end{table}

	\subsection{Implementation Details}
	
	\subsubsection{Segmentation Network}
	We initially train the S-Net network using the Helen dataset \cite{helen2012} which contains $2000$ clean images with corresponding semantic labels provided by Smith \etal\cite{Smith2013}. S-Net is trained using the cross entropy loss with the Adam optimizer and the learning rate is set equal to $0.0002$. We train S-Net for $60000$ iterations on clean images.  As can be seen from Fig.~\ref{Fig:Seg} and Table~\ref{Table:Seg}, S-Net trained on clean images only is not able to produce good results on the blurry images. Thus, we fine-tune S-Net using the blurry images and the corresponding semantic maps generated using the Helen dataset \cite{helen2012}. In this case, S-Net is fine-tuned for $30000$ iterations with the learning rate of $0.00001$. We test our S-Net network on $330$ clean Test images provided by the Helen dataset as well as $8000$ blurry images. Results are shown in Fig.~\ref{Fig:Seg} and Table~\ref{Table:Seg} in terms of F-scores.  As can be seen from these results, our method is able to produce reasonable segmentation results from blurry images.
	
	\setlength{\tabcolsep}{5pt}
	\def\arraystretch{1.1}
	\begin{table}[htp!]
		\resizebox{0.48\textwidth}{!}{
			\begin{tabular}{c|c|c|c}
				\hline
				\multirow{2}{*}{Class} & Clean Images & \multicolumn{2}{c}{Blurry Images}
				\\ \cline{2-4}
				& \begin{tabular}[c]{@{}c@{}}S-Net trained on\\  clean images\end{tabular} & \begin{tabular}[c]{@{}c@{}}S-Net trained on\\  clean images\end{tabular} & \begin{tabular}[c]{@{}c@{}}S-Net fine-tuned w/ \\ blurry images\end{tabular} \\ \hline
				class-1 & 0.910 & 0.889 & 0.901 \\ \hline
				class-2 & 0.903 & 0.877 & 0.884 \\ \hline
				class-3 & 0.856 & 0.775 & 0.814 \\ \hline
				class-4 & 0.625 & 0.538 & 0.566 \\ \hline
			\end{tabular}
		}
		\vspace{0.5mm}
		\caption{F-score values of semantic masks for clean and blurry images from the Helen dataset.}
		\label{Table:Seg}
	\end{table}

	\subsubsection{UMSN Network}
	Training images for UMSN are generated using $2000$ images from the Helen dataset \cite{helen2012}, and randomly selected $25000$ images from the CelebA dataset \cite{liu2015faceattributes}. We generate $25000$ blur kernels sizes ranging from $13 \times 13$ to $29 \times 29$, using 3D camera trajectories \cite{Boracchi2012}. Patches of size $128 \times 128$ are exacted from those images and convolved with $25000$ blur kernels randomly to generated about 1.7 million pairs of clean-blurry data. We added Gaussian noise with $\sigma=0.03$ to the blurry images.
	
\begin{figure}[htp!]
		\centering
		\includegraphics[width=0.11\textwidth,height= 0.11\textwidth]{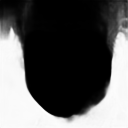}
		\includegraphics[width=0.11\textwidth,height= 0.11\textwidth]{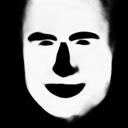}
		\includegraphics[width=0.11\textwidth,height= 0.11\textwidth]{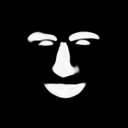}
		\includegraphics[width=0.11\textwidth,height= 0.11\textwidth]{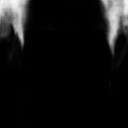}\\
		\vskip2pt
		\includegraphics[width=0.11\textwidth,height= 0.11\textwidth]{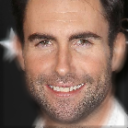}
		\includegraphics[width=0.11\textwidth,height= 0.11\textwidth]{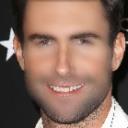}
		\includegraphics[width=0.11\textwidth,height= 0.11\textwidth]{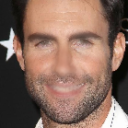}
		\includegraphics[width=0.11\textwidth,height= 0.11\textwidth]{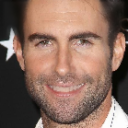}\\
		\vskip2pt
		\includegraphics[width=0.11\textwidth,height= 0.11\textwidth]{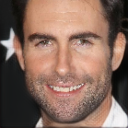}
		\includegraphics[width=0.11\textwidth,height= 0.11\textwidth]{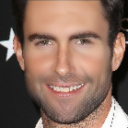}
		\includegraphics[width=0.11\textwidth,height= 0.11\textwidth]{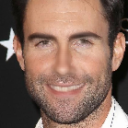}
		\includegraphics[width=0.11\textwidth,height= 0.11\textwidth]{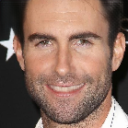} \\
		(a)\hskip40pt(b)\hskip50pt(c)\hskip40pt(d)\\
		\caption{First row: semantic masks. Second row: blurry images. Third row: deblurred images using first stage semantic network for classes (a) $m_1$, (b) $m_2$ , (c) $m_3$ (d) $m_4$. Note that first stage semantic network is combination of F-Net-$i$, ResBlock and Conv$3\times 3$.}
		\label{Fig:exp8}
	\end{figure}

	\paragraph{First Stage Semantic Network}
	Every first stage semantic network is trained with clean-blurry paired data where only the corresponding class components of the face images are blurred.  The first stage semantic network is trained with a combination of the L1-norm and the perceptual losses using the Adam optimizer with learning rate of $0.0002$. The first stage network is trained for $50,000$ iterations.  To illustrate what kind of outputs are observed from the first stage semantic networks, in Fig.~\ref{Fig:exp8} we display semantic masks (first row), the corresponding blurry images (second row) and the final estimated deblurred outputs from first stage semantic networks (last row).  As it can be seen from the last row, first stage semantic networks are able to remove most of the blur from corresponding class in the image. Note that, F-Net-$i$'s from these first stage semantic networks are used along with B-Net to construct UMSN, and trained to perform deblurring face image.

	\paragraph{Overall UMSN Network}
	UMSN is trained using $\mathcal{L}_{total}$ with the Adam optimizer and  batchsize of 16.  The learning rate is set equal to $0.0002$. Note that, semantic maps $\{{m_i}\}_{i=1}^{4}$ of the ground truth clean images in $\mathcal{L}_{total}$ are generated using S-Net, since we do not have the actual semantic maps for them. $\lambda$ and $\lambda_1$ values are set equal to $0.01$ and $0.0002$, respectively. UMSN is trained for $0.1$ million iterations.

\subsection{Evaluation}
\paragraph{PSNR and SSIM} We compare the performance of our method with the following state-of-the-art algorithms: MAP-based methods \cite{Krishnan2011,Xu2013,hqdeblurring_siggraph2008,ChoLee2009,Zhong2013}, face deblurring exemplar-based method \cite{PanECCV14}, and CNN-based methods \cite{Nah_2017_CVPR,ZiyiDeep,kupyn2019deblurgan}. Results are shown in Table~\ref{Table:Comp}. As can be seen from this table, UMSN outperforms state-of-the-art methods including the methods that make use of semantic maps for image deblurring \cite{ZiyiDeep}. Furthermore, we evaluate UMSN's performance in reconstructing individual semantic classes against \cite{ZiyiDeep}. As it can be seen from the Table~\ref{Table:class}, our method's performance in reconstructing individual classes is better than the sate-of-the-art method \cite{ZiyiDeep}. In addition, we compare the performance of our method against CNN-based face deblurring method that uses post processing \cite{song-ijcv19-FHD} on the  PubFig dataset consisting of 192 images released by the authors of \cite{song-ijcv19-FHD}. The PSNR and SSIM values achieved by \cite{song-ijcv19-FHD} are $30.21$ and $0.84$, whereas our method achieved $30.95$ and $0.87$, respectively. 
		
	\setlength{\tabcolsep}{5pt}
	\def\arraystretch{1.2}
	\begin{table}[t!]
		\resizebox{0.5\textwidth}{!}{
			\begin{tabular}{l|c|c|c|c}
				\hline
				\multirow{2}{*}{Deblurring Method} & \multicolumn{2}{c|}{Helen} & \multicolumn{2}{c}{CelebA} \\ \cline{2-5}
				& PSNR & SSIM & PSNR & SSIM \\ \hline
				Krishnan \etal\cite{Krishnan2011} (CVPR'11) & 19.30 & 0.670 & 18.38 & 0.672 \\ \hline
				Pan \etal\cite{PanECCV14} (ECCV 2014) & 20.93 & 0.727 & 18.59 & 0.677 \\ \hline
				Shan  \etal\cite{hqdeblurring_siggraph2008} (SIGGRAPH'08) & 19.57 & 0.670 & 18.43 & 0.644          \\ \hline
				Xu  \etal\cite{Xu2013} (CVPR'13) & 20.11 & 0.711 & 18.93 & 0.685 \\ \hline
				Cho \etal\cite{ChoLee2009} (SIGGRAPH'09) & 16.82 & 0.574 & 13.03 & 0.445 \\ \hline
				Zhong  \etal\cite{Zhong2013} (CVPR'13) & 16.41 & 0.614 & 17.26 & 0.695 \\ \hline
				Nah  \etal\cite{Nah_2017_CVPR} (CVPR'17) & 24.12 & 0.823 & 22.43 & 0.832 \\ \hline
				\begin{tabular}[c]{@{}l@{}}Shen \etal\cite{ZiyiDeep} (CVPR'18) w/GAN\end{tabular} & 25.58 & 0.861 & 24.34 & 0.860 \\ \hline
				\begin{tabular}[c]{@{}l@{}}Shen \etal\cite{ZiyiDeep} (CVPR'18) \end{tabular} & 25.99 & 0.871 & 25.05 & 0.879 \\ \hline
				Kupyn  \etal\cite{kupyn2019deblurgan} (ICCV'19) & 26.45 & 0.880 & 25.42 & 0.884 \\ \hline
				UMSN (ours w/$\mathcal{L}_{total}$) & \textbf{27.75} & \textbf{0.897} & \textbf{26.62} & \textbf{0.908} \\ \hline
			\end{tabular}
		}
		\vspace{3pt}
		\caption{PSNR and SSIM comparision of UMSN against state-of-the-art methods.
			\label{Table:Comp}
		}
	\end{table}

\paragraph{VGG-Face distance} We compare the performance of our method with the following state-of-the-art algorithms using feature distance ( i.e, $L2-$norm distance between output feature map from network) between deblurred image and the ground truth image. We use the outputs of \textit{pool5} layer from the VGG-Face~\cite{Parkhi15} to compute the VGG-Face distance (~$d_{VGG}$). $d_{VGG}$ is a better perceptual metric to compare as it computes the $L2-$norm distance in the feature space using VGG-Face~\cite{Parkhi15} network. Table~\ref{dVGG} clearly shows that our method UMSN outperforms the existing state of the art methods using the perceptual measures like $d_{VGG}$.
	 
	\setlength{\tabcolsep}{5pt}
	\begin{table}[htp!]
		\center
		\resizebox{0.4\textwidth}{!}{
			\begin{tabular}{l|c|c}
				\hline
				\multirow{2}{*}{Deblurring Method}                                                                                                         & Helen               & CelebA     \\ \cline{2-3} 
				& $d_{VGG}$ & $d_{VGG}$ \\ \hline
				Xu  \etal\cite{Xu2013} (CVPR'13)                                                                           & 10.15               & 11.38      \\ \hline
				Zhong  \etal\cite{Zhong2013} (CVPR'13)                                                                                             & 12.28               & 14.72      \\ \hline
				Nah  \etal \cite{Nah_2017_CVPR} (CVPR'17)                                                                               & 7.46                & 8.92       \\ \hline
				\begin{tabular}[c]{@{}l@{}}Shen \etal\cite{ZiyiDeep} (CVPR'18) \end{tabular} & 5.87                & 6.49       \\ \hline
				Kupyn  \etal\cite{kupyn2019deblurgan} (ICCV'19) & 4.91  & 5.84     \\ \hline
				UMSN (ours w/$\mathcal{L}_{total}$)                                                                                                        & \textbf{2.69}  & \textbf{2.85}      \\ \hline
			\end{tabular}
		}
		\vspace{3pt}
		\caption{comparision of UMSN against state-of-the-art methods using  distance of feature from VGG-Face(~$d_{VGG}$), lower is better \label{dVGG}}
	\end{table}
	\begin{figure*}[htp!]
		\begin{center}
			\centering
			\includegraphics[width=0.135\textwidth]{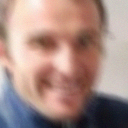}
			\includegraphics[width=0.135\textwidth]{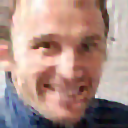}
			\includegraphics[width=0.135\textwidth]{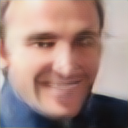}
			\includegraphics[width=0.135\textwidth]{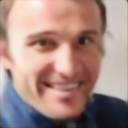}
			\includegraphics[width=0.135\textwidth]{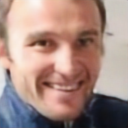}
			\includegraphics[width=0.135\textwidth]{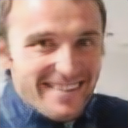}
			\includegraphics[width=0.135\textwidth]{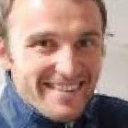}\\
			\includegraphics[width=0.135\textwidth]{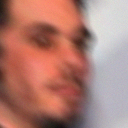}
			\includegraphics[width=0.135\textwidth]{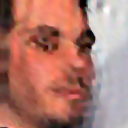}
			\includegraphics[width=0.135\textwidth]{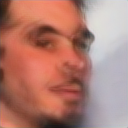}
			\includegraphics[width=0.135\textwidth]{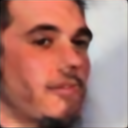}
			\includegraphics[width=0.135\textwidth]{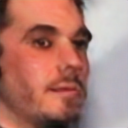}
			\includegraphics[width=0.135\textwidth]{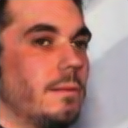}
			\includegraphics[width=0.135\textwidth]{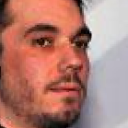}\\
			\includegraphics[width=0.135\textwidth]{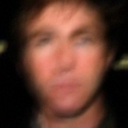}
			\includegraphics[width=0.135\textwidth]{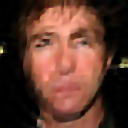}
			\includegraphics[width=0.135\textwidth]{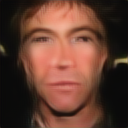}
			\includegraphics[width=0.135\textwidth]{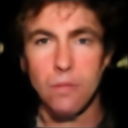}
			\includegraphics[width=0.135\textwidth]{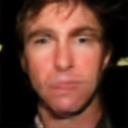}
			\includegraphics[width=0.135\textwidth]{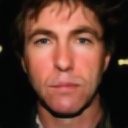}
			\includegraphics[width=0.135\textwidth]{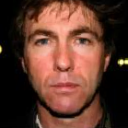}\\
			\includegraphics[width=0.135\textwidth]{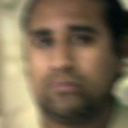}
			\includegraphics[width=0.135\textwidth]{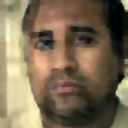}
			\includegraphics[width=0.135\textwidth]{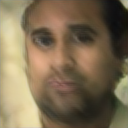}
			\includegraphics[width=0.135\textwidth]{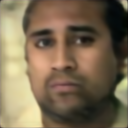}
			\includegraphics[width=0.135\textwidth]{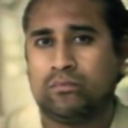}
			\includegraphics[width=0.135\textwidth]{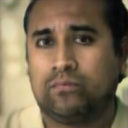}
			\includegraphics[width=0.135\textwidth]{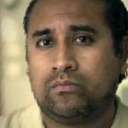}\\
			\includegraphics[width=0.135\textwidth]{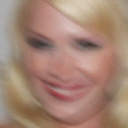}
			\includegraphics[width=0.135\textwidth]{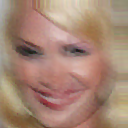}
			\includegraphics[width=0.135\textwidth]{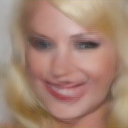}
			\includegraphics[width=0.135\textwidth]{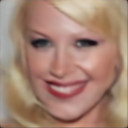}
			\includegraphics[width=0.135\textwidth]{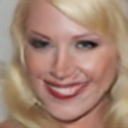}
			\includegraphics[width=0.135\textwidth]{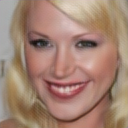}
			\includegraphics[width=0.135\textwidth]{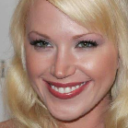}\\
			\includegraphics[width=0.135\textwidth]{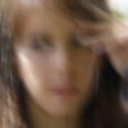}
			\includegraphics[width=0.135\textwidth]{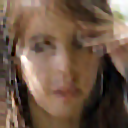}
			\includegraphics[width=0.135\textwidth]{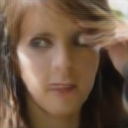}
			\includegraphics[width=0.135\textwidth]{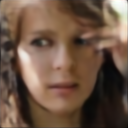}
			\includegraphics[width=0.135\textwidth]{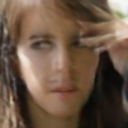}
			\includegraphics[width=0.135\textwidth]{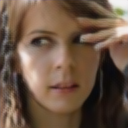}
			\includegraphics[width=0.135\textwidth]{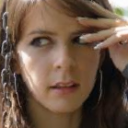}\\
			\begin{flushleft}
			\hskip25ptBlurry\hskip35ptXu \etal\cite{Xu2013}\hskip15ptZhong \etal\cite{Zhong2013}\hskip15ptShen \etal\cite{ZiyiDeep}\hskip8pt Kupyn \etal\cite{kupyn2019deblurgan} \hskip20ptOurs \hskip35ptGround Truth \\
			\hskip25ptImage\hskip45pt(CVPR'13)\hskip25pt(CVPR'13)\hskip30pt(CVPR'18)\hskip25pt(ICCV'19)\hskip40ptFinal \hskip45pt Image \\
			\vspace{2mm}
			\end{flushleft}
			\caption{Sample results from the Helen and CelebA datasets. As can be seen from this figure, MAP-based methods \cite{Zhong2013,Xu2013} are adding some artifacts on the final deblurred images especially near the eyes, nose and mouth regions.  Deblurring \cite{kupyn2019deblurgan} without prior produced smooth output face images with artifacts, this can be seen in the fifth column corresponding to \cite{kupyn2019deblurgan}.  Global prior-based method \cite{ZiyiDeep} produces very smooth outputs and some parts of the deblurred images are deformed.  This can be clearly seen by looking at the reconstructions in the  fourth column corresponding to \cite{ZiyiDeep}, where the right eye is added in the face image, the fingers are removed, the mustache is removed, and the mouth is deformed. On the other hand by using class-specific multi-stream networks and $\mathcal{L}_{total}$, our method is able to reconstruct all the classes of a face perfectly and produces sharper face images. For example, UMSN is able to reconstruct eyes, nose, mustache and mouth clearly.  Additionally UMSN does not add parts like eyes when they are occluded (last row). Note that the method proposed in \cite{ZiyiDeep} uses a generative adversarial network (GAN) to reconstruct sharper images. In contrast, our method is able to reconstruct sharper images even without using a GAN.}
			\label{Fig:exp6}
		\end{center}
		
	\end{figure*}
	\begin{figure*}[htp!]
		\begin{center}
			\centering
			\includegraphics[width=0.135\textwidth]{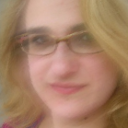}
			\includegraphics[width=0.135\textwidth]{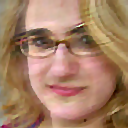}
			\includegraphics[width=0.135\textwidth]{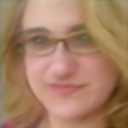}
			\includegraphics[width=0.135\textwidth]{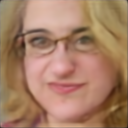}
			\includegraphics[width=0.135\textwidth]{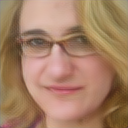}
			\includegraphics[width=0.135\textwidth]{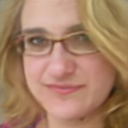}\\
			\vskip2pt
			\includegraphics[width=0.135\textwidth]{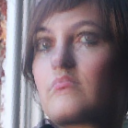}
			\includegraphics[width=0.135\textwidth]{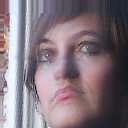}
			\includegraphics[width=0.135\textwidth]{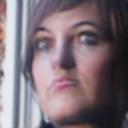}
			\includegraphics[width=0.135\textwidth]{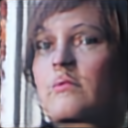}
			\includegraphics[width=0.135\textwidth]{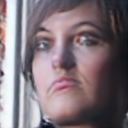}
			\includegraphics[width=0.135\textwidth]{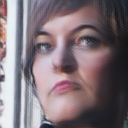}\\
			\vskip2pt
			\includegraphics[width=0.135\textwidth]{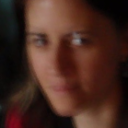}
			\includegraphics[width=0.135\textwidth]{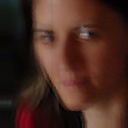}
			\includegraphics[width=0.135\textwidth]{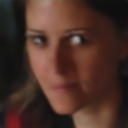}
			\includegraphics[width=0.135\textwidth]{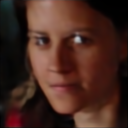}
			\includegraphics[width=0.135\textwidth]{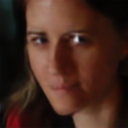}
			\includegraphics[width=0.135\textwidth]{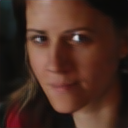}\\
			\vskip2pt
			\includegraphics[width=0.135\textwidth]{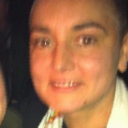}
			\includegraphics[width=0.135\textwidth]{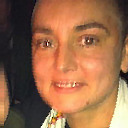}
			\includegraphics[width=0.135\textwidth]{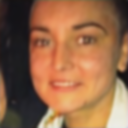}
			\includegraphics[width=0.135\textwidth]{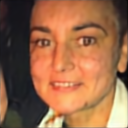}
			\includegraphics[width=0.135\textwidth]{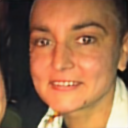}
			\includegraphics[width=0.135\textwidth]{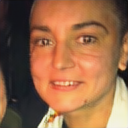}\\
			\vskip2pt
			\includegraphics[width=0.135\textwidth]{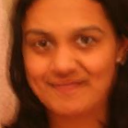}
			\includegraphics[width=0.135\textwidth]{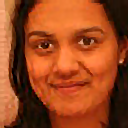}
			\includegraphics[width=0.135\textwidth]{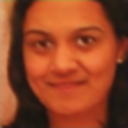}
			\includegraphics[width=0.135\textwidth]{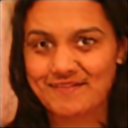}
			\includegraphics[width=0.135\textwidth]{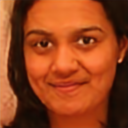}
			\includegraphics[width=0.135\textwidth]{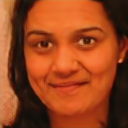}\\
			\vskip2pt
			\includegraphics[width=0.135\textwidth]{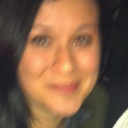}
			\includegraphics[width=0.135\textwidth]{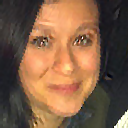}
			\includegraphics[width=0.135\textwidth]{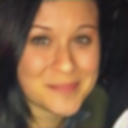}
			\includegraphics[width=0.135\textwidth]{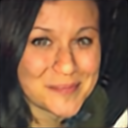}
			\includegraphics[width=0.135\textwidth]{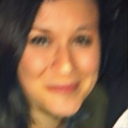}
			\includegraphics[width=0.135\textwidth]{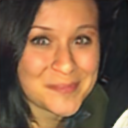}\\
			\begin{flushleft}
				\hskip65ptBlurry\hskip30ptXu \etal\cite{Xu2013}\hskip15ptZhong \etal\cite{Zhong2013}\hskip15ptShen \etal\cite{ZiyiDeep}\hskip8pt Kupyn \etal\cite{kupyn2019deblurgan} \hskip20ptOurs  \\
				\hskip65ptImage\hskip45pt(CVPR'13)\hskip25pt(CVPR'13)\hskip30pt(CVPR'18)\hskip25pt(ICCV'19)\hskip40ptFinal \\
				\vspace{2mm}
			\end{flushleft}
			\vspace{2mm}
			\caption{Sample results on real blurry images.}
			\label{Fig:exp_real}
		\end{center}
	\end{figure*}
	\begin{figure*}[htp!]
		\begin{center}
			\centering
			\includegraphics[width=0.12\textwidth,height=0.13\textwidth]{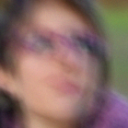}
			\includegraphics[width=0.12\textwidth,height=0.13\textwidth]{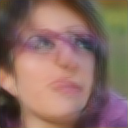}
			\includegraphics[width=0.12\textwidth,height=0.13\textwidth]{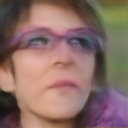}
			\includegraphics[width=0.12\textwidth,height=0.13\textwidth]{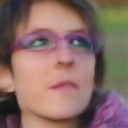}
			\includegraphics[width=0.12\textwidth,height=0.13\textwidth]{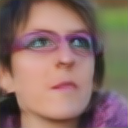}
			\includegraphics[width=0.12\textwidth,height=0.13\textwidth]{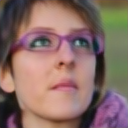}
			\includegraphics[width=0.12\textwidth,height=0.13\textwidth]{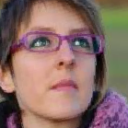}\\
			\vskip2pt
			\includegraphics[width=0.12\textwidth,height=0.13\textwidth]{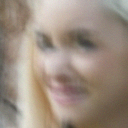}
			\includegraphics[width=0.12\textwidth,height=0.13\textwidth]{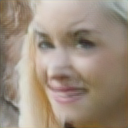}
			\includegraphics[width=0.12\textwidth,height=0.13\textwidth]{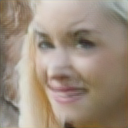}
			\includegraphics[width=0.12\textwidth,height=0.13\textwidth]{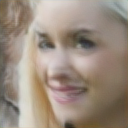}
			\includegraphics[width=0.12\textwidth,height=0.13\textwidth]{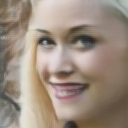}
			\includegraphics[width=0.12\textwidth,height=0.13\textwidth]{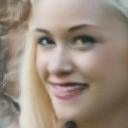}
			\includegraphics[width=0.12\textwidth,height=0.13\textwidth]{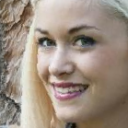}\\
			(a)\hskip55pt(b)\hskip55pt(c)\hskip50pt(d)\hskip50pt(e)\hskip55pt(f)\hskip55pt(g)\\
			\caption{Ablation study. (a) Blurry image, (b) B-Net, (c) B-Net + semantic maps, (d) B-Net + semantic maps + NRL, (e) UMSN without $\mathcal{L}_{c}$, (f) UMSN (g) ground Truth.}
			\label{Fig:exp4}
		\end{center}
		
	\end{figure*}
	\paragraph{Face recognition}
	In order to show the significance of different face deblurring methods, we perform face recognition on the deblurred images.   We use the CelebA dataset to perform this comparison, where we use $100$ different identities as the probe set, and for each identity we select 9 additional clean face images as the gallery set. $8000$ blurry images are generated using the 100 images in the probe set. Given a blurry image, the deblurred image is computed using the corresponding deblurring algorithm. The most similar face for this deblurred image is selected from the gallery set to check whether they belong to same identity or not. 
	
	To perform face recognition we use OpenFace toolbox~\cite{amos2016openface} to compute the similarity distance between the deblurred image and all the gallery images. We select the Top-K nearest matches for each deblurred image to compute the accuracy of the deblurring method. Table~\ref{detect} clearly shows that the deblurred images by our method have better recognition accuracies than the other state-of-the-art methods.  This experiment clearly shows that our method is able to retain the important parts of a face while performing deblurring.  This in turn helps in achieving better face recognition compared to the other methods.

	\begin{table}[htp!]
		\center
		\resizebox{0.45\textwidth}{!}{
		\begin{tabular}{l|c|c|c}
			\hline
			Deblurring Method & Top-1 & Top-3 & Top-5 \\ \hline
			Clear images      & 71\%  & 84\%  & 89\%  \\ \hline
			Blurred images    & 31\%  & 46\%  & 53\%  \\ \hline
			Pan \etal\cite{PanECCV14} (ECCV 2014)               & 44\%  & 57\%  & 64\%  \\ \hline
			Xu  \etal\cite{Xu2013} (CVPR'13)                & 43\%  & 57\%  & 64\%  \\ \hline
			Zhong  \etal\cite{Zhong2013} (CVPR'13)             & 30\%  & 44\%  & 51\%  \\ \hline
			Nah  \etal\cite{Nah_2017_CVPR} (CVPR'17)               & 42\%  & 59\%  & 65\%  \\ \hline
			\begin{tabular}[c]{@{}l@{}}Shen \etal\cite{ZiyiDeep} (CVPR'18) \end{tabular}              & 54\%  & 68\%  & 74\%  \\ \hline
			Kupyn  \etal\cite{kupyn2019deblurgan} (ICCV'19) & 58\% & 70\% & 77\% \\ \hline
			UMSN (ours w/$\mathcal{L}_{total}$)              & \textbf{64}\%  & \textbf{75}\%  & \textbf{83}\% \\ \hline
		\end{tabular}
		}
		\vspace{3pt}
		\caption{Top-1, Top-3 and Top-5 face recognition accuracies on the CelebA dataset. \label{detect}}
	\end{table}
	
	\paragraph{Qualitative Results}
	The qualitative performance of different methods on several images from the Helen and CelebA test images are shown in  Fig.~\ref{Fig:exp6}. We can clearly observe that UMSN produces sharp images without any artifacts when compared to the other methods \cite{Xu2013,ZiyiDeep,Zhong2013}.
	
	In addition, we conducted experiments on 100 real blurry images provided by the authors of \cite{Xu2013,ZiyiDeep,Zhong2013}. Results corresponding to different methods are shown in Fig.~\ref{Fig:exp_real}.   As can be seen from this figure, UMSN produces sharper and clear images when compared to the state-of-the-art methods. For example, methods \cite{Xu2013,Zhong2013} produce results which contain artifacts or  blurry images. Shen \etal\cite{ZiyiDeep}, Kupyn \etal\cite{kupyn2019deblurgan} are not able reconstruct eyes, nose and mouth as shown in the fourth and fifth columns of Fig.~\ref{Fig:exp_real}, respectively. On the other hand, eyes, nose, and mouth regions are clearly visible in the images corresponding to the UMSN method. 
	
	\subsection{Ablation Study}
	We conduct experiments on two datasets to study the performance contribution of each component in UMSN.  We start with the base network (B-Net) and progressively add components to establish their significance to estimate the final deblurred image.  The corresponding results are shown in Table~\ref{Table:Abl}. It is important to note that B-Net and UMSN are configured to have the same number of learnable parameters by setting appropriate number of output channels in the intermediate convolutional layers of the ResBlock. The ablation study demonstrates that each of the components improve the quality of the deblurred image. In particular, the introduction of the multi-stream architecture improved the performance by~$0.6$dB.  When UMSN is trained using $\mathcal{L}_{total}$ the performance improved by an additional $0.75$dB. Finally, the combination of all the components of UMSN produce the best results with $2.5$dB gain compared to B-Net.

	\setlength{\tabcolsep}{5pt}
	\def\arraystretch{1.2}
	\begin{table}[h!]
		\center
		\resizebox{0.4\textwidth}{!}{
			\begin{tabular}{l|c|c|c|c}
				\hline
				\multirow{2}{*}{Deblurring Method} & \multicolumn{2}{c|}{Helen} & \multicolumn{2}{c}{CelebA} \\ \cline{2-5}
				& PSNR & SSIM & PSNR & SSIM \\ \hline
				baseline (B-Net)  & 25.12 & 0.82 & 24.38 & 0.81\\ \hline
				+ semantic maps & 25.55 & 0.84 & 24.72 & 0.82 \\ \hline
				+ nested residuals & 26.08 & 0.85 & 25.26 & 0.84 \\ \hline
				UMSN without $\mathcal{L}_{c}$ & 26.96 & 0.87 & 25.90 & 0.87  \\ \hline
				UMSN & \textbf{27.75} & \textbf{0.90} & \textbf{26.62} & \textbf{0.91} \\ \hline
			\end{tabular}
		}
		\vspace{5pt}
		\caption{PSNR and SSIM results of the ablation study.}
		\label{Table:Abl}
	\end{table}

\begin{figure}[t!]
		\centering
		\includegraphics[width=0.11\textwidth]{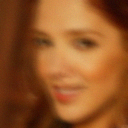}
		\includegraphics[width=0.11\textwidth]{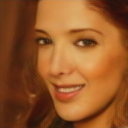}
		\includegraphics[width=0.11\textwidth]{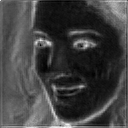}
		\includegraphics[width=0.11\textwidth]{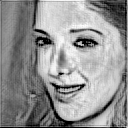}\\
		\includegraphics[width=0.11\textwidth]{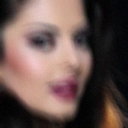}
		\includegraphics[width=0.11\textwidth]{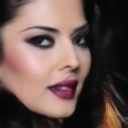}
		\includegraphics[width=0.11\textwidth]{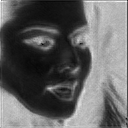}
		\includegraphics[width=0.11\textwidth]{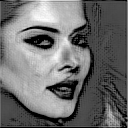}\\
		(a)\hskip40pt(b)\hskip50pt(c)\hskip40pt(d)\\
		\caption{(a)  Blurry image. (b) UMSN. (c) Intermediate residual feature map learned for the eyes and the hair. (d) Intermediate residual feature map learned  for the face skin in NRL.}
		\label{Fig:exp5}
	\end{figure}

	Sample reconstructions corresponding to the ablation study using the Helen and CelebA datasets are shown in Fig.~\ref{Fig:exp4}.  As can be seen from this figure, B-Net and B-Net+semantic maps produce reconstructions that are still blurry and they fail to reconstruct the facial parts like eyes, nose and mouth well. On the other hand, B-Net+semantic maps+ NRL is able to improve the reconstruction quality but visually it is still not comparable to the state-of-the-art method. As shown in Fig. \ref{Fig:exp5}, NRL is able to estimate the class-specific residual feature maps, which are further used for estimating the final deblurred image. From Fig. \ref{Fig:exp5} we can see what different residual feature maps are learning for different classes.  In particular, the use of F-Nets helps UMSN to produce qualitatively good results, where  all the parts of a face are clearly visible. Furthermore, training UMSN using $\mathcal{L}_{total}$ results in sharper images.
	
\begin{figure}[h!]
		\centering
		\includegraphics[width=0.11\textwidth,height= 0.12\textwidth]{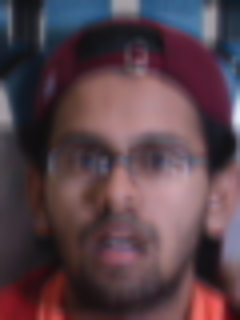}
		\includegraphics[width=0.11\textwidth,height= 0.12\textwidth]{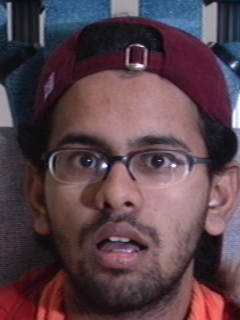}
		\includegraphics[width=0.11\textwidth,height= 0.12\textwidth]{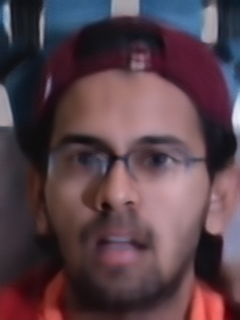}
		\includegraphics[width=0.11\textwidth,height= 0.12\textwidth]{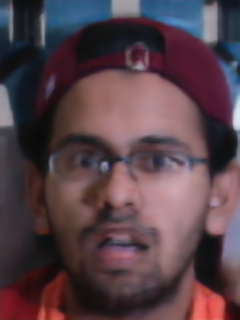}\\
		\includegraphics[width=0.11\textwidth,height= 0.12\textwidth]{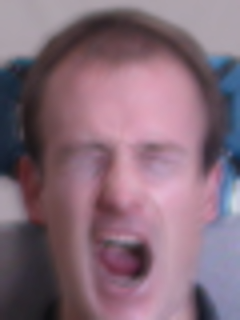}
		\includegraphics[width=0.11\textwidth,height= 0.12\textwidth]{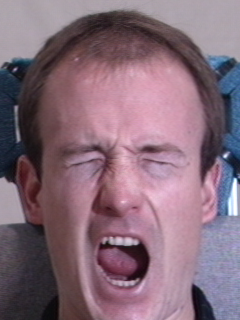}
		\includegraphics[width=0.11\textwidth,height= 0.12\textwidth]{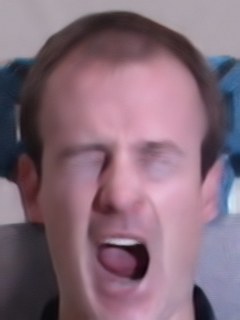}
		\includegraphics[width=0.11\textwidth,height= 0.12\textwidth]{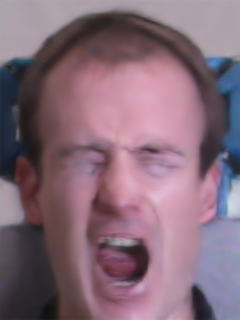}\\
		(a)\hskip40pt(b)\hskip50pt(c)\hskip40pt(d)\\
		\caption{Sample results from the PubFig dataset. (a) Blurry image. (b) Ground Truth. (c) Song \etal\cite{song-ijcv19-FHD}(IJCV'19). (d) UMSN.}
		\label{Fig:exp7}
	\end{figure}

	We also compare our method qualitatively with post processing-based method \cite{song-ijcv19-FHD} on the PubFig dataset. As can be seen from the results shown in Fig.~\ref{Fig:exp7}, even after applying some post processing, \cite{song-ijcv19-FHD} tends to produce results that are smooth. In comparison, UMSN is able to produce sharper and high-quality images without any post-processing.

\section{Conclusion}\label{sec:con}
We proposed a new method, called Uncertainty guided Multi-stream Semantic Network (UMSN), to address the single image blind deblurring of face image problem that entails the use of facial sementic information. In our approach, we introduced a different way to use the ouput features from the sub-networks which are trained for the individual semantic class-wise deblurring. Additionally, we introduced novel techniques such as nested residual learning, and class-based confidence loss to improve the performance of UMSN. In comparison to the state-of-the-art single image blind deblurring methods, the proposed approach is able to achieve significant improvements when evaluated on three popular face datasets.

\ifCLASSOPTIONcaptionsoff
  \newpage
\fi



%

\appendices
\section{UMSN network sequence}

The BaseNetwork is similar to the first stage semantic network as shown in Fig.~\ref{Fig:F-Net} with the following sequence of layers,\\
ResBlock(3,64)-Avgpool-ResBlock(64,64)-ResBlock(64,64)-ResBlock(64,64)-ResBlock(64,64)-ResBlock(64,64)-ResBlock(64,64)-Upsample-ResBlock(64,16)-Conv$~3\times~3$(16,3).\\
F-Net-$i$ is a a sequence of five ResBlocks with dense connections,\\
ResBlock(3,16)-Avgpool-ResBlock(16,16)-ResBlock(16,16)-ResBlock(16,16)-ResBlock(16,8)\\
where ResBlock$(m,n)$ indicates $m$ input channels and $n$ output channels for ResBlock. 

F-Net-$i$'s are joined to form the first stage of UMSN, further these outputs re concatenated to output of first layer of Base Network(B-Net) for constructing the UMSN network. Thus UMSN is constructed by combining F-Net-$i$'s and B-Net, where F-Net-$i$'s acts as first stage of UMSN, and B-Net acts as second stage of UMSN.
\begin{figure*}[htp!]
	\centering
	\includegraphics[width=0.15\textwidth]{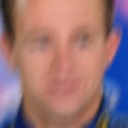}
	\includegraphics[width=0.15\textwidth]{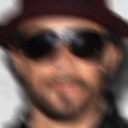}
	\includegraphics[width=0.15\textwidth]{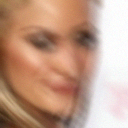}
	\includegraphics[width=0.15\textwidth]{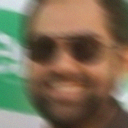}
	\includegraphics[width=0.15\textwidth]{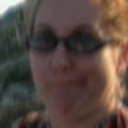}
	\includegraphics[width=0.15\textwidth]{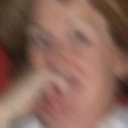}\\ \vskip2pt
	\includegraphics[width=0.15\textwidth]{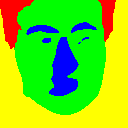}
	\includegraphics[width=0.15\textwidth]{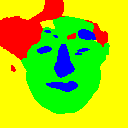}
	\includegraphics[width=0.15\textwidth]{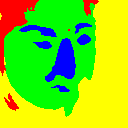}
	\includegraphics[width=0.15\textwidth]{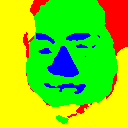}
	\includegraphics[width=0.15\textwidth]{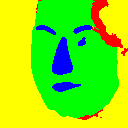}
	\includegraphics[width=0.15\textwidth]{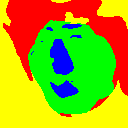}
	\\ \vskip2pt
	\includegraphics[width=0.15\textwidth]{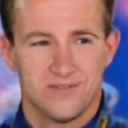}
	\includegraphics[width=0.15\textwidth]{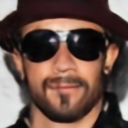}
	\includegraphics[width=0.15\textwidth]{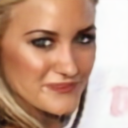}
	\includegraphics[width=0.15\textwidth]{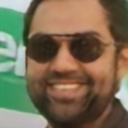}
	\includegraphics[width=0.15\textwidth]{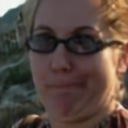}
	\includegraphics[width=0.15\textwidth]{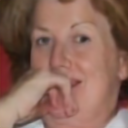}
	\\ \vskip2pt
	\includegraphics[width=0.15\textwidth]{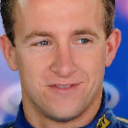}
	\includegraphics[width=0.15\textwidth]{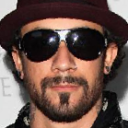}
	\includegraphics[width=0.15\textwidth]{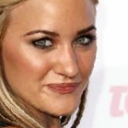}
	\includegraphics[width=0.15\textwidth]{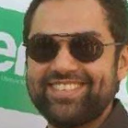}
	\includegraphics[width=0.15\textwidth]{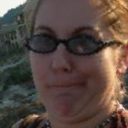}
	\includegraphics[width=0.15\textwidth]{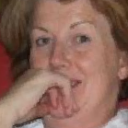}
	\caption{Semantic maps generated by S-Net on a blurry image from the Helen dataset. First row contains Blurry images. Second rows consists of corresponding Semantic masks obtained from S-Net. Third consists of deblurred images using UMSN. Fourth row contains ground-truth images.}
	\label{Fig:Oc_res}
	
\end{figure*}

\section{Confidence scores}
\begin{figure}[htp!]
	\centering
	\includegraphics[width=0.115\textwidth]{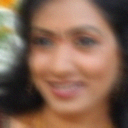}
	\includegraphics[width=0.115\textwidth]{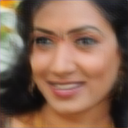} 
	\includegraphics[width=0.115\textwidth]{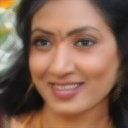}
	\includegraphics[width=0.115\textwidth]{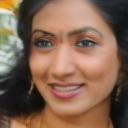}\\
	\vskip2pt
	\includegraphics[width=0.115\textwidth]{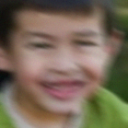}
	\includegraphics[width=0.115\textwidth]{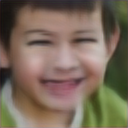} 
	\includegraphics[width=0.115\textwidth]{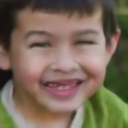}
	\includegraphics[width=0.115\textwidth]{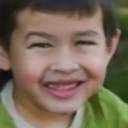}\\
	(a)\hskip40pt(b)\hskip50pt(c)\hskip40pt(d)\\
	\caption{Sample deblurring results corresponding to UMSN at different time instances of training. (a) Blurry image. (b) Trained for 3000 iterations, (c) 50,000 iterations, and (d) 0.1 million iterations.
	}
	\label{Fig:exp_sup}
\end{figure}

\begin{figure}[htp!]
	\centering
	\includegraphics[width=0.115\textwidth]{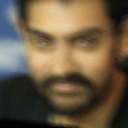}
	\includegraphics[width=0.115\textwidth]{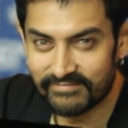} 
	\includegraphics[width=0.115\textwidth]{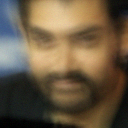}
	\includegraphics[width=0.115\textwidth]{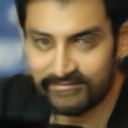}\\
	\vskip2pt
	\includegraphics[width=0.115\textwidth]{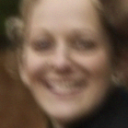}
	\includegraphics[width=0.115\textwidth]{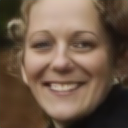} 
	\includegraphics[width=0.115\textwidth]{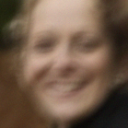}
	\includegraphics[width=0.115\textwidth]{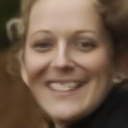}\\
	(a)\hskip40pt(b)\hskip50pt(c)\hskip40pt(d)\\
	\caption{Sample deblurring results of UMSN for different kernel sizes  $17\times 17$ and $27\times 27$ for two different identiies. (a) Blurry images with kernel size $17 \times 17$. (b) Deblurred images using UMSN. (c) Blurry image with kernel size $27 \times 27$. (d) Deblurred image using UMSN.}
	\label{Fig:exp_sup2}
\end{figure}

In this section, we show some example test outputs of UMSN (trained with $\mathcal{L}_{total}$) at different time instances of training, and their corresponding confidence values to show how confidence measure is helping the network. Confidence measure helps the UMSN in deblurring different parts of the face. For example, as shown in the Fig.~\ref{Fig:exp_sup} and Table \ref{Table:2}, the outputs of UMSN after 50,000 iterations are blurry around eyes, nose, and mouth regions which is reflected in low confidence scores for $C_3$ when compared to the other classes. Eventually, UMSN learns to reconstruct eyes, nose and mouth regions which is reflected in high confidence scores for $C_3$.  This can be seen from Figure \ref{Fig:exp_sup} where the output deblurred images are sharp with eyes, nose and mouth regions reconstructed perfectly. Table~\ref{Table:cnf_1} and Fig.~\ref{Fig:exp_sup2} show the confidence scores for different classes and outputs of UMSN for two different identities from the test dataset blurred with kernel sizes of $17\times 17$ and $27\times 27$. 

\begin{table}[htp!]
	\begin{center}
	\begin{tabular}{l|l|l|l|l|l}
		\hline
		Blurry Image            & Iteration & $C_1$  & $C_2$  & $C_3$  & $C_4$  \\ \hline
		\multirow{3}{*}{Image1} & 3000      & 0.573 & 0.576 & 0.610 & 0.584 \\ \cline{2-6} 
		& 50000     & 0.676 & 0.902 & 0.789 & 0.847 \\ \cline{2-6} 
		& 100000    & 0.858 & 0.917 & 0.989 & 0.962 \\ \hline
		\multirow{3}{*}{Image2} & 3000      & 0.519 & 0.528 & 0.493 & 0.611 \\ \cline{2-6} 
		& 50000     & 0.728 & 0.883 & 0.817 & 0.872 \\ \cline{2-6} 
		& 100000    & 0.896 & 0.919 & 0.966 & 0.944 \\ \hline
	\end{tabular}
	\vspace{2mm}
	\caption{Confidence values corresponding to Image1 and Image2 in first row and second row images in Fig.~\ref{Fig:exp_sup}, respectively.}
	\label{Table:2}
\end{center}
\end{table}

\begin{table}[htp!]
	\begin{center}
	\begin{tabular}{c|l|l|l|l|l}
		\hline
		Clean image Identity            & kernel & $C_1$  & $C_2$  & $C_3$  & $C_4$  \\ \hline
		\multirow{2}{*}{Identity1} & $17 \times 17$      & 0.923 & 0.948 & 0.913 &  0.973 \\ \cline{2-6}  
		& $27 \times 27$    & 0.871 & 0.832 & 0.848 & 0.945 \\ \hline
		\multirow{2}{*}{Identity2} & $17 \times 17$      & 0.874 & 0.931 & 0.942 &  0.945 \\ \cline{2-6}  
		& $27 \times 27$    & 0.817 & 0.844 & 0.869 & 0.880 \\ \hline
	\end{tabular}
	\vspace{2mm}
	\caption{Confidence values for the blurry images generated using Identity1 and Identity2 corresponding to first row and second row images in Fig.~\ref{Fig:exp_sup2}, respectively.}
	\label{Table:cnf_1}
\end{center}
\end{table}

\section{Additional results}
Extreme blur and occlusions by object like sun glasses or other body parts such as hands can cause S-Net to produce poor quality segmentations as shown in the second row of Fig.~\ref{Fig:Oc_res}. Even when the S-Net fails to produce accurate segmentation masks, UMSN is able to produce visually pleasing results as shown in the third row of Fig.~\ref{Fig:Oc_res}. This is mainly due to the concatenation of features from F-Net-$i$ with the feature from the first layer of B-Net.  By doing this, features extracted and reconstructed for each class in corresponding F-Net helps the UMSN to deblur the face image even if the face parsing does not work well. Note that the first stage semantic networks are initially trained on the images where only the corresponding class is blurred and remaining are clean.   Thus even if the class mask is extremely corrupted, the F-Net can extract the corresponding class features and helps the UMSN to produce visually pleasing images as shown in Fig.~\ref{Fig:Oc_res}.

\begin{figure}[htp!]
	\centering
	\includegraphics[width=0.15\textwidth]{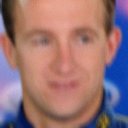}
	\includegraphics[width=0.15\textwidth]{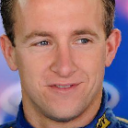}
	\includegraphics[width=0.15\textwidth]{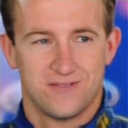}\\ \vskip2pt
	\includegraphics[width=0.15\textwidth]{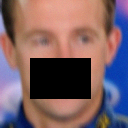}
	\includegraphics[width=0.15\textwidth]{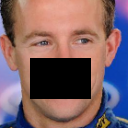}
	\includegraphics[width=0.15\textwidth]{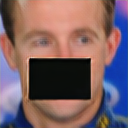}\\ \vskip2pt
	(a) \hskip55pt (b) \hskip55pt (c)  
	\caption{First row corresponds to the images without occlusion and second row corresponds to the images with occlusion. (a) Blurry images. (b)  Ground-turth clean images. (c) Deblurred images using the proposed UMSN network.}
	\label{Fig:fail1}
	
\end{figure}

\begin{figure}[htp!]
	\centering
	\includegraphics[width=0.15\textwidth]{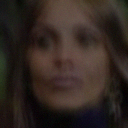}
	\includegraphics[width=0.15\textwidth]{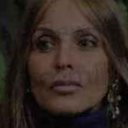}
	\includegraphics[width=0.15\textwidth]{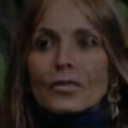}\\ \vskip2pt
	(a) \hskip55pt (b) \hskip55pt (c)  
	\caption{(a) Blurry image. (b) Ground-turth clean image. (c)  Deblurred images using the proposed UMSN network.}
	\label{Fig:fail2}
	
\end{figure}

Figure~\ref{Fig:fail1} and Figure~\ref{Fig:fail2} show cases in which the face images are occluded and taken in low-light conditions, respectively. In such conditions, the proposed method fails to produce better quality images.

\section*{Acknowledgment}

This research is based upon work supported by the Office
of  the  Director  of  National  Intelligence  (ODNI),  Intelligence  Advanced  Research  Projects  Activity  (IARPA),  via IARPA  R\&D  Contract  No.  2019-19022600002.  The  views
and  conclusions  contained  herein  are  those  of  the  authors
and should not be interpreted as necessarily representing the
official policies or endorsements, either expressed or implied,
of  the  ODNI,  IARPA,  or  the  U.S.  Government.  The  U.S.
Government is authorized to reproduce and distribute reprints
for  Governmental  purposes  notwithstanding  any  copyright
annotation  thereon.

\bibliography{Deblur_TIP19}

\begin{thebibliography}{10}
\providecommand{\url}[1]{#1}
\csname url@samestyle\endcsname
\providecommand{\newblock}{\relax}
\providecommand{\bibinfo}[2]{#2}
\providecommand{\BIBentrySTDinterwordspacing}{\spaceskip=0pt\relax}
\providecommand{\BIBentryALTinterwordstretchfactor}{4}
\providecommand{\BIBentryALTinterwordspacing}{\spaceskip=\fontdimen2\font plus
\BIBentryALTinterwordstretchfactor\fontdimen3\font minus
  \fontdimen4\font\relax}
\providecommand{\BIBforeignlanguage}[2]{{%
\expandafter\ifx\csname l@#1\endcsname\relax
\typeout{** WARNING: IEEEtran.bst: No hyphenation pattern has been}%
\typeout{** loaded for the language `#1'. Using the pattern for}%
\typeout{** the default language instead.}%
\else
\language=\csname l@#1\endcsname
\fi
#2}}
\providecommand{\BIBdecl}{\relax}
\BIBdecl

\bibitem{Fergus06removingcamera}
R.~Fergus, B.~Singh, A.~Hertzmann, S.~T. Roweis, and W.~T. Freeman, ``Removing
  camera shake from a single photograph,'' \emph{ACM transactions on graphics
  (TOG)}, vol.~25, no.~3, pp. 787--794, 2006.

\bibitem{Schuler2013}
C.~J. Schuler, H.~Christopher~Burger, S.~Harmeling, and B.~Scholkopf, ``A
  machine learning approach for non-blind image deconvolution,'' in
  \emph{Proceedings of the IEEE Conference on Computer Vision and Pattern
  Recognition}, 2013, pp. 1067--1074.

\bibitem{ShearDeconv}
V.~M. {Patel}, G.~R. {Easley}, and D.~M. {Healy}, ``Shearlet-based
  deconvolution,'' \emph{IEEE Transactions on Image Processing}, vol.~18,
  no.~12, pp. 2673--2685, Dec 2009.

\bibitem{ManifoldDeblur}
J.~{Ni}, P.~{Turaga}, V.~M. {Patel}, and R.~{Chellappa}, ``Example-driven
  manifold priors for image deconvolution,'' \emph{IEEE Transactions on Image
  Processing}, vol.~20, no.~11, pp. 3086--3096, Nov 2011.

\bibitem{Ren2016}
W.~Ren, X.~Cao, J.~Pan, X.~Guo, W.~Zuo, and M.-H. Yang, ``Image deblurring via
  enhanced low-rank prior,'' \emph{IEEE Transactions on Image Processing},
  vol.~25, no.~7, pp. 3426--3437, 2016.

\bibitem{patchdeblur_iccp2013}
L.~Sun, S.~Cho, J.~Wang, and J.~Hays, ``Edge-based blur kernel estimation using
  patch priors,'' in \emph{IEEE International Conference on Computational
  Photography (ICCP)}.\hskip 1em plus 0.5em minus 0.4em\relax IEEE, 2013, pp.
  1--8.

\bibitem{Nimisha17}
T.~M. Nimisha, A.~Kumar~Singh, and A.~N. Rajagopalan, ``Blur-invariant deep
  learning for blind-deblurring,'' in \emph{Proceedings of the IEEE
  International Conference on Computer Vision}, 2017, pp. 4752--4760.

\bibitem{ayan2016}
L.~Xu, J.~S. Ren, C.~Liu, and J.~Jia, ``Deep convolutional neural network for
  image deconvolution,'' in \emph{Advances in neural information processing
  systems}, 2014, pp. 1790--1798.

\bibitem{Nah_2017_CVPR}
S.~Nah, T.~Hyun~Kim, and K.~Mu~Lee, ``Deep multi-scale convolutional neural
  network for dynamic scene deblurring,'' in \emph{Proceedings of the IEEE
  Conference on Computer Vision and Pattern Recognition}, 2017, pp. 3883--3891.

\bibitem{Zhang_2018_CVPR}
S.~Zhang, X.~Shen, Z.~Lin, R.~M{\v{e}}ch, J.~P. Costeira, and J.~M. Moura,
  ``Learning to understand image blur,'' in \emph{Proceedings of the IEEE
  Conference on Computer Vision and Pattern Recognition}, 2018, pp. 6586--6595.

\bibitem{ZiyiDeep}
Z.~Shen, W.-S. Lai, T.~Xu, J.~Kautz, and M.-H. Yang, ``Deep semantic face
  deblurring,'' \emph{Proceedings of the IEEE Conference on Computer Vision and
  Pattern Recognition}, pp. 8260--8269, 2018.

\bibitem{song-ijcv19-FHD}
Y.~Song, J.~Zhang, L.~Gong, S.~He, L.~Bao, J.~Pan, Q.~Yang, and M.-H. Yang,
  ``Joint face hallucination and deblurring via structure generation and detail
  enhancement,'' \emph{International Journal of Computer Vision}, vol. 127, no.
  6-7, pp. 785--800, 2019.

\bibitem{helen2012}
V.~Le, J.~Brandt, Z.~Lin, L.~Bourdev, and T.~S. Huang, ``Interactive facial
  feature localization,'' in \emph{European conference on computer
  vision}.\hskip 1em plus 0.5em minus 0.4em\relax Springer, 2012, pp. 679--692.

\bibitem{liu2015faceattributes}
Z.~Liu, P.~Luo, X.~Wang, and X.~Tang, ``Deep learning face attributes in the
  wild,'' in \emph{Proceedings of the IEEE international conference on computer
  vision}, 2015, pp. 3730--3738.

\bibitem{Xu2013}
L.~Xu, S.~Zheng, and J.~Jia, ``Unnatural l0 sparse representation for natural
  image deblurring,'' \emph{Proceedings of the IEEE conference on computer
  vision and pattern recognition}, pp. 1107--1114, 2013.

\bibitem{Krishnan2011}
D.~Krishnan, T.~Tay, and R.~Fergus, ``Blind deconvolution using a normalized
  sparsity measure,'' \emph{CVPR 2011}, pp. 233--240, 2011.

\bibitem{Pan_2016_CVPR}
J.~Pan, D.~Sun, H.~Pfister, and M.-H. Yang, ``Blind image deblurring using dark
  channel prior,'' in \emph{Proceedings of the IEEE Conference on Computer
  Vision and Pattern Recognition}, 2016, pp. 1628--1636.

\bibitem{Kruse2017}
J.~Kruse, C.~Rother, and U.~Schmidt, ``Learning to push the limits of efficient
  fft-based image deconvolution,'' in \emph{Proceedings of the IEEE
  International Conference on Computer Vision}, 2017, pp. 4586--4594.

\bibitem{Schmidt2013}
U.~Schmidt, C.~Rother, S.~Nowozin, J.~Jancsary, and S.~Roth, ``Discriminative
  non-blind deblurring,'' in \emph{Proceedings of the IEEE Conference on
  Computer Vision and Pattern Recognition}, 2013, pp. 604--611.

\bibitem{Schmidt2014}
U.~Schmidt and S.~Roth, ``Shrinkage fields for effective image restoration,''
  in \emph{Proceedings of the IEEE Conference on Computer Vision and Pattern
  Recognition}, 2014, pp. 2774--2781.

\bibitem{vasu2018non}
S.~Vasu, V.~Reddy~Maligireddy, and A.~Rajagopalan, ``Non-blind deblurring:
  Handling kernel uncertainty with cnns,'' in \emph{Proceedings of the IEEE
  Conference on Computer Vision and Pattern Recognition}, 2018, pp. 3272--3281.

\bibitem{Schuler_NIPS14}
C.~J. Schuler, M.~Hirsch, S.~Harmeling, and B.~Sch{\"o}lkopf, ``Learning to
  deblur,'' vol.~38, no.~7.\hskip 1em plus 0.5em minus 0.4em\relax IEEE, 2015,
  pp. 1439--1451.

\bibitem{Xu2014}
A.~Chakrabarti, ``Deep convolutional neural network for image deconvolution,''
  2014.

\bibitem{kupyn2019deblurgan}
O.~Kupyn, T.~Martyniuk, J.~Wu, and Z.~Wang, ``Deblurgan-v2: Deblurring
  (orders-of-magnitude) faster and better,'' in \emph{Proceedings of the IEEE
  International Conference on Computer Vision}, 2019, pp. 8878--8887.

\bibitem{TsaiSLSY16}
Y.~Tsai, X.~Shen, Z.~Lin, K.~Sunkavalli, and M.~Yang, ``Sky is not the limit:
  semantic-aware sky replacement,'' \emph{{ACM} Trans. Graph.}, vol.~35, no.~4,
  2016.

\bibitem{EigenKF13}
D.~Eigen, D.~Krishnan, and R.~Fergus, ``Restoring an image taken through a
  window covered with dirt or rain,'' in \emph{Proceedings of the IEEE
  International Conference on Computer Vision}, 2013, pp. 633--640.

\bibitem{SvobodaHMZ16}
P.~Svoboda, M.~Hradi{\v{s}}, L.~Mar{\v{s}}{\'\i}k, and P.~Zemc{\'\i}k, ``Cnn
  for license plate motion deblurring,'' in \emph{2016 IEEE International
  Conference on Image Processing (ICIP)}.\hskip 1em plus 0.5em minus
  0.4em\relax IEEE, 2016, pp. 3832--3836.

\bibitem{PanHSY14}
J.~Pan, Z.~Hu, Z.~Su, and M.-H. Yang, ``Deblurring text images via
  l0-regularized intensity and gradient prior,'' in \emph{Proceedings of the
  IEEE Conference on Computer Vision and Pattern Recognition}, 2014, pp.
  2901--2908.

\bibitem{BakerK02}
S.~Baker and T.~Kanade, ``Limits on super-resolution and how to break them,''
  \emph{{IEEE} Trans. Pattern Anal. Mach. Intell.}, vol.~24, no.~9, 2002.

\bibitem{PanECCV14}
J.~Pan, Z.~Hu, Z.~Su, and M.-H. Yang, ``Deblurring face images with
  exemplars,'' pp. 47--62, 2014.

\bibitem{JoshiMAK10}
N.~Joshi, W.~Matusik, E.~H. Adelson, and D.~J. Kriegman, ``Personal photo
  enhancement using example images,'' \emph{{ACM} Trans. Graph.}, vol.~29,
  no.~2, 2010.

\bibitem{bulat2018super}
A.~Bulat and G.~Tzimiropoulos, ``Super-fan: Integrated facial landmark
  localization and super-resolution of real-world low resolution faces in
  arbitrary poses with gans,'' in \emph{Proceedings of the IEEE Conference on
  Computer Vision and Pattern Recognition}, 2018, pp. 109--117.

\bibitem{chen2018gated}
D.~Chen, M.~He, Q.~Fan, J.~Liao, L.~Zhang, D.~Hou, L.~Yuan, and G.~Hua, ``Gated
  context aggregation network for image dehazing and deraining,'' pp.
  1375--1383, 2019.

\bibitem{yu2018face}
X.~Yu, B.~Fernando, B.~Ghanem, F.~Porikli, and R.~Hartley, ``Face
  super-resolution guided by facial component heatmaps,'' in \emph{Proceedings
  of the European Conference on Computer Vision (ECCV)}, 2018, pp. 217--233.

\bibitem{ren2019face}
W.~Ren, J.~Yang, S.~Deng, D.~Wipf, X.~Cao, and X.~Tong, ``Face video deblurring
  using 3d facial priors,'' in \emph{Proceedings of the IEEE International
  Conference on Computer Vision}, 2019, pp. 9388--9397.

\bibitem{lu2019unsupervised}
B.~Lu, J.-C. Chen, and R.~Chellappa, ``Unsupervised domain-specific deblurring
  via disentangled representations,'' in \emph{Proceedings of the IEEE
  Conference on Computer Vision and Pattern Recognition}, 2019, pp.
  10\,225--10\,234.

\bibitem{Instance2016}
D.~Ulyanov, A.~Vedaldi, and V.~Lempitsky, ``Instance normalization: The missing
  ingredient for fast stylization,'' \emph{arXiv preprint arXiv:1607.08022},
  2016.

\bibitem{Authors15b}
\BIBentryALTinterwordspacing
O.~Ronneberger, P.Fischer, and T.~Brox", ``U-net: Convolutional networks for
  biomedical image segmentation",'' in \emph{Medical Image Computing and
  Computer-Assisted Intervention (MICCAI)"}, ser. "LNCS", vol. "9351", "2015",
  pp. "234--241". [Online]. Available:
  \url{"http://lmb.informatik.uni-freiburg.de/Publications/2015/RFB15a"}
\BIBentrySTDinterwordspacing

\bibitem{huang2017densely}
G.~Huang, Z.~Liu, L.~Van Der~Maaten, and K.~Q. Weinberger, ``Densely connected
  convolutional networks,'' in \emph{Proceedings of the IEEE conference on
  computer vision and pattern recognition}, 2017, pp. 4700--4708.

\bibitem{wang2018smoothed}
Z.~Wang and S.~Ji, ``Smoothed dilated convolutions for improved dense
  prediction,'' in \emph{Proceedings of the 24th ACM SIGKDD International
  Conference on Knowledge Discovery \& Data Mining}.\hskip 1em plus 0.5em minus
  0.4em\relax ACM, 2018, pp. 2486--2495.

\bibitem{Johnson2016Perceptual}
J.~Johnson, A.~Alahi, and L.~Fei-Fei, ``Perceptual losses for real-time style
  transfer and super-resolution,'' \emph{ECCV}, 2016.

\bibitem{zhang2017multistyle}
H.~Zhang and K.~Dana, ``Multi-style generative network for real-time
  transfer,'' \emph{arXiv preprint arXiv:1703.06953}, 2017.

\bibitem{Ledig2017PhotoRealisticSI}
C.~Ledig, L.~Theis, F.~Husz{\'a}r, J.~Caballero, A.~Cunningham, A.~Acosta,
  A.~Aitken, A.~Tejani, J.~Totz, Z.~Wang \emph{et~al.}, ``Photo-realistic
  single image super-resolution using a generative adversarial network,'' pp.
  4681--4690, 2017.

\bibitem{Parkhi15}
O.~M. Parkhi, A.~Vedaldi, and A.~Zisserman, ``Deep face recognition,'' in
  \emph{British Machine Vision Conference}, 2015.

\bibitem{SSIM}
Z.~Wang, A.~C. Bovik, H.~R. Sheikh, and E.~P. Simoncelli, ``Image quality
  assessment: from error visibility to structural similarity,'' \emph{IEEE
  Transactions on Image Processing}, vol.~13, no.~4, pp. 600--612, April 2004.

\bibitem{Smith2013}
B.~M. Smith, L.~Zhang, J.~Brandt, Z.~Lin, and J.~Yang, ``Exemplar-based face
  parsing,'' in \emph{Proceedings of the IEEE Conference on Computer Vision and
  Pattern Recognition}, 2013, pp. 3484--3491.

\bibitem{Boracchi2012}
G.~Boracchi and A.~Foi, ``Modeling the performance of image restoration from
  motion blur,'' \emph{IEEE Transactions on Image Processing}, vol.~21, no.~8,
  pp. 3502--3517, 2012.

\bibitem{hqdeblurring_siggraph2008}
Q.~Shan, J.~Jia, and A.~Agarwala, ``High-quality motion deblurring from a
  single image,'' \emph{Acm transactions on graphics (tog)}, vol.~27, no.~3,
  p.~73, 2008.

\bibitem{ChoLee2009}
S.~Cho and S.~Lee, ``Fast motion deblurring,'' \emph{ACM Transactions on
  graphics (TOG)}, vol.~28, no.~5, p. 145, 2009.

\bibitem{Zhong2013}
L.~Zhong, S.~Cho, D.~Metaxas, S.~Paris, and J.~Wang, ``Handling noise in single
  image deblurring using directional filters,'' in \emph{Proceedings of the
  IEEE Conference on Computer Vision and Pattern Recognition}, 2013, pp.
  612--619.

\bibitem{amos2016openface}
B.~Amos, B.~Ludwiczuk, and M.~Satyanarayanan, ``Openface: A general-purpose
  face recognition library with mobile applications,'' CMU-CS-16-118, CMU
  School of Computer Science, Tech. Rep., 2016.

\end{thebibliography}
\bibliographystyle{IEEEtran}
\vskip-150pt
\begin{IEEEbiography}
	[{\includegraphics[width=1in,height=1in,clip,keepaspectratio]{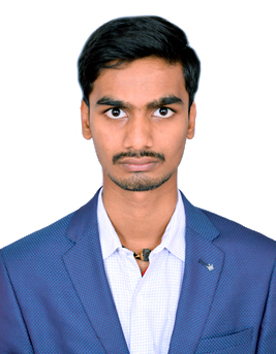}}]%
	{Rajeev Yasarla}
	is PhD student in the Department of Electrical and Computer Engineering (ECE) at Johns Hopkins University. Prior to joining Hopkins, he graduated from IIT Madras with Bachelor's and Master's degree in Electrical Engineering. His research interests include deep learning based  image restoration (like de-raining, deblurring, and atomspheric turbulence distortion removal), object detection and medical image segmentation. 
	
\end{IEEEbiography}
\vskip-125pt
\begin{IEEEbiography}
	[{\includegraphics[width=1in,height=1.5in,clip,keepaspectratio]{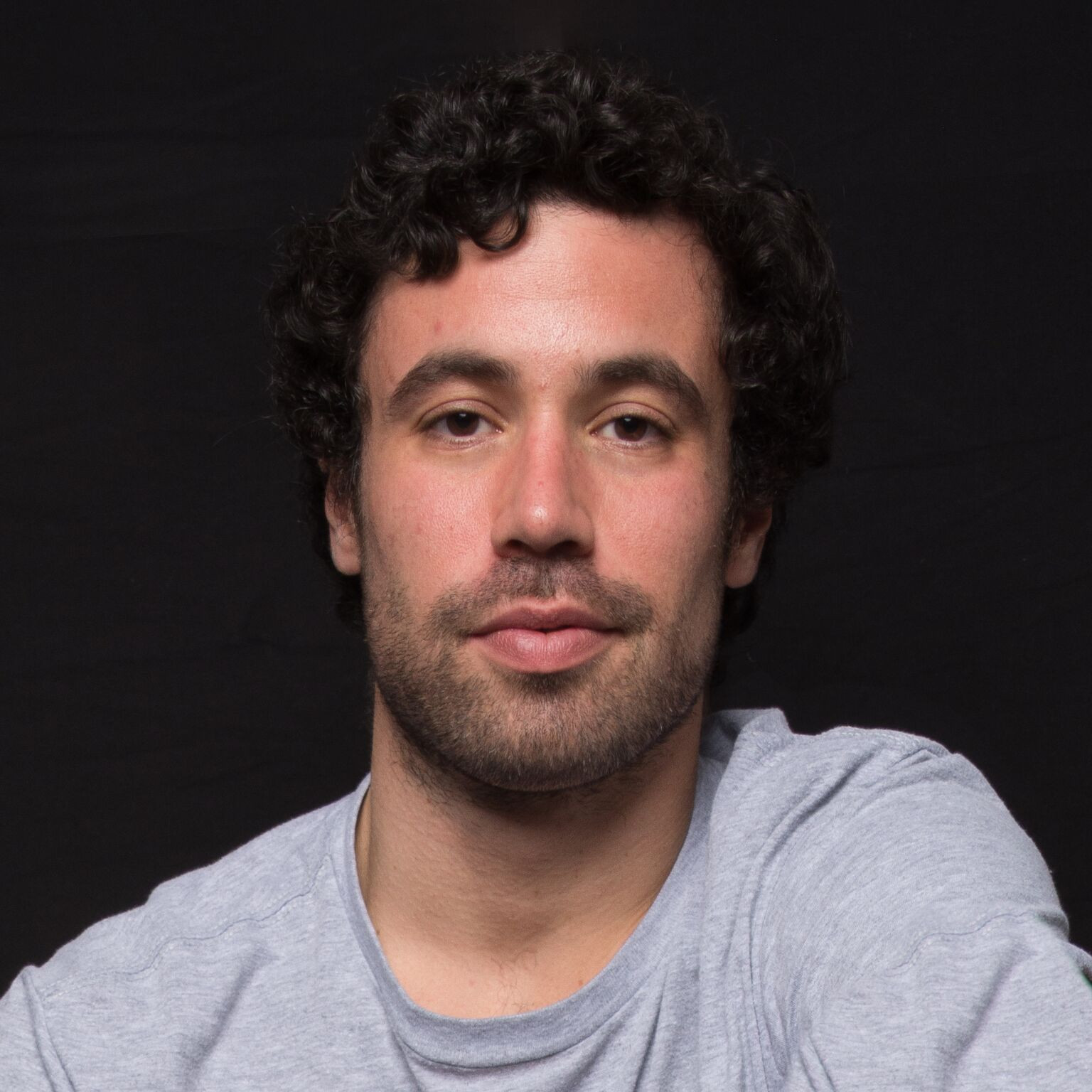}}]%
	{Federico Perazzi}
	 is a Research Scientist at Adobe Research. Prior to that, Federico was a post-doctoral researcher at Disney Research Zurich, Switzerland. Federico received a degree in Computer Science in 2008, a degree in Electrical Engineering in 2008, an M.Sc. in Computer Science and a Ph.D. from the Swiss Federal Institute of Technology (ETHZ), the M.Sc. in Entertainment Technology in 2010 from Carnegie Mellon University. 
	
\end{IEEEbiography}
\vskip-125pt
\begin{IEEEbiography}
	[{\includegraphics[width=1.25in,height=1.25in,clip,keepaspectratio]{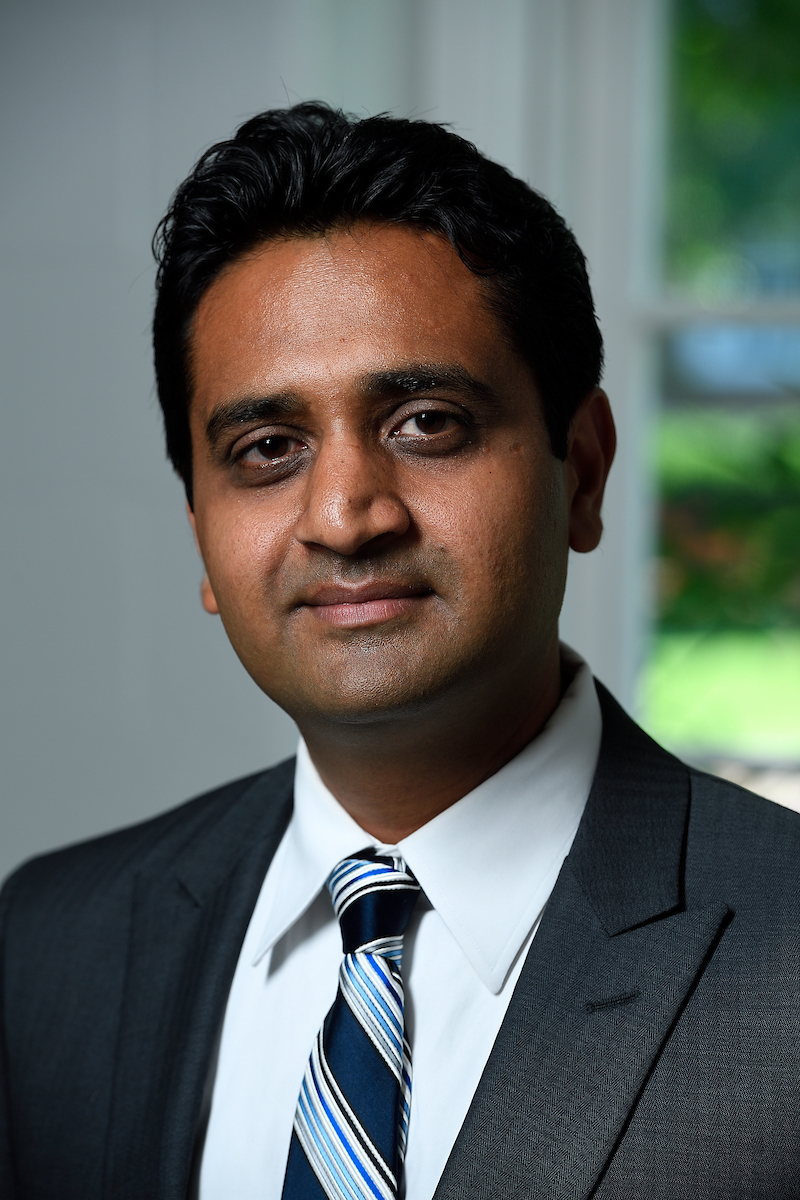}}]%
	{Vishal M. Patel}
	\text{[SM'15]} is an Assistant Professor in the Department of Electrical and Computer Engineering (ECE) at Johns Hopkins University. Prior to joining Hopkins, he was an A. Walter Tyson Assistant Professor in the Department of ECE at Rutgers University and a member of the research faculty at the University of Maryland Institute for Advanced Computer Studies (UMIACS). He completed his Ph.D. in Electrical Engineering from the University of Maryland, College Park, MD, in 2010.  His  current  research interests include signal processing, computer vision,and  pattern  recognition  with  applications  in  biometrics  and  imaging.    He has received a number of awards including the 2016 ONR Young Investigator Award, the 2016 Jimmy Lin Award for Invention, A. Walter Tyson Assistant Professorship Award, Best Paper Award at IEEE AVSS 2017 $\&$ 2019, Best Paper Award at IEEE BTAS 2015, Honorable Mention Paper Award at IAPR ICB 2018, two Best Student Paper Awards at IAPR ICPR 2018, and Best Poster Awards at BTAS 2015 and 2016. He is an Associate Editor of the IEEE Signal Processing Magazine, IEEE Biometrics Compendium, and serves on the Machine Learning for Signal Processing (MLSP) Committee of the IEEE Signal Processing Society. He is a member of Eta Kappa Nu, Pi Mu Epsilon, and Phi Beta Kappa.
	
\end{IEEEbiography}

\end{document}